\documentclass{article}

\usepackage[preprint]{neurips_2025}

\usepackage[utf8]{inputenc} 
\usepackage[T1]{fontenc}    
\usepackage{hyperref}       
\usepackage{url}            
\usepackage{booktabs}       
\usepackage{amsfonts}       
\usepackage{nicefrac}       
\usepackage{microtype}      
\usepackage{xcolor}  
\usepackage{soul}
\usepackage{tcolorbox}

\usepackage{graphicx}   
\usepackage{subcaption} 
\usepackage{array}      
\usepackage{colortbl}   

\usepackage[table]{xcolor}
\definecolor{pelicanmain}{HTML}{FBE6B2}
\definecolor{pelicanlight}{HTML}{FFF3D6}
\definecolor{myred}{RGB}{220, 20, 60}

\usepackage{multirow}
\usepackage{multicol}
\usepackage{wrapfig}
\usepackage{algorithm}
\usepackage{algpseudocode}
\usepackage{changepage}

\usepackage{amssymb, amsmath}
\usepackage{dsfont}
\usepackage{cleveref}

\usepackage{mathrsfs}
\usepackage{enumitem}
\usepackage{caption}
\usepackage{bbm}
\usepackage{marvosym}


\title{Dream4ACT: A Shared Visual Action Interface for Multi-Embodiment Video-Action Modeling}

\author{
  \parbox{0.95\textwidth}{
    \centering
    Xiangyu Zhu$^{1,2}$, Jin Xu$^{1}$, Yue Guo$^{1,3}$,
    Xin Wu$^{1,4}$, Yifan Sun$^{1,5}$, Xiancong Ren$^{1}$, \\
    Jianxin Sun$^{1}$,
    Yong Dai$^{1}$$^{\dagger}$,
    Xiaozhu Ju$^{1}$\protect\textsuperscript{\Letter}
    \\[0.6em]
    $^{1}$Beijing Humanoid Robot Innovation Center,
    $^{2}$Beijing Institute of Technology, \\
    $^{3}$Harbin Institute of Technology, Shenzhen
    $^{4}$The University of Hong Kong \\
    $^{5}$China University of Mining \& Technology, Beijing
    \\[0.4em]
    $^{\dagger}$ Project leader, \protect\textsuperscript{\Letter} Corresponding author \\ [0.5em]
    \href{https://dream4act.github.io/}{https://dream4act.github.io/}
  }
}

\begin{document}

\maketitle

\begin{abstract}

\noindent\makebox[\linewidth][r]{%
\begin{minipage}{\textwidth}
\begin{tcolorbox}[
    colback=gray!15,
    colframe=white,
    boxrule=0pt,
    arc=2mm,
    left=4mm,
    right=4mm,
    width=\linewidth
]
Video generation models (VGMs) offer strong spatiotemporal priors for embodied observation-action modeling. However, joint-space action vectors lack explicit image-space structure and vary in dimensionality and semantics across embodiments, making it challenging to directly leverage the rich spatiotemporal priors of VGMs. End-effector visualizations provide an alternative but do not specify the full articulated configuration needed for robot execution. We present \textbf{Dream4ACT}, a world model built for joint video-action modeling across embodiments. To unify action representations across embodiments, we introduce a shared visual action interface, called \emph{action views}, which render target joint configurations from four prescribed virtual cameras using URDF-based forward kinematics. This shared visual representation preserves embodiment-specific articulated geometry while allowing observation and action sequences to share a video autoencoder and diffusion transformer. Through masked flow-matching, our model supports forward dynamics, inverse dynamics, and joint observation-action generation within a single jointly trained model by varying which future sequences are corrupted. To recover executable action sequences from predicted action views, we propose a training-free, URDF-constrained multiview recovery mechanism, without a learned embodiment-specific decoder. Dream4ACT achieves an average success rate of 88.98\% on RoboTwin~2.0 and an overall score of 65.66 on TriWorldBench, supporting effective closed-loop manipulation and competitive action-conditioned multiview prediction through the visual action interface.
\end{tcolorbox}
\end{minipage}%
}

\end{abstract}

\section{Introduction}
\label{sec:introduction}

Video-action world models offer a promising foundation for embodied intelligence by connecting visual prediction with robot control. A general model should support two complementary directions: predicting future observations under candidate controls and inferring controls consistent with observed or desired future observations. Large-scale video pretraining provides spatiotemporal priors~\cite{wan2025wan}, motivating growing interest in adapting video models for robot prediction and control~\cite{cheang2024gr,li2026causal}.

Recent works use video models either to provide visual plans or predictive features for a separate controller~\cite{du2023learning,hu2024video}, or to jointly predict observations and actions, often through modality-specific heads~\cite{li2025unifiedvideoactionmodel,li2026causal}. Joint-space vectors lack explicit spatial structure and vary across embodiments. Shared policies support heterogeneous action spaces~\cite{team2024octo,liu2025rdt}, while latent-frame encodings reuse video generation for action prediction~\cite{kim2026cosmos}. Visual interfaces further represent end-effector quantities as images~\cite{li2026spatialvamspatialawaremultiviewvideodiffusion,zhen2026action}, but do not explicitly encode the full arm configuration. Executing these end-effector commands still requires an embodiment-specific mapping to joint targets, potentially with singular solutions or no solutions.

Our key insight is to represent target articulated configurations as visual content modeled alongside RGB observations. For joint-controlled manipulation, we seek a shared visual interface that retains arm and gripper configuration information, supports joint training across embodiments, and permits recovery of executable joint targets.

\begin{figure*}[t]
    \centering
    \includegraphics[width=0.95\linewidth]{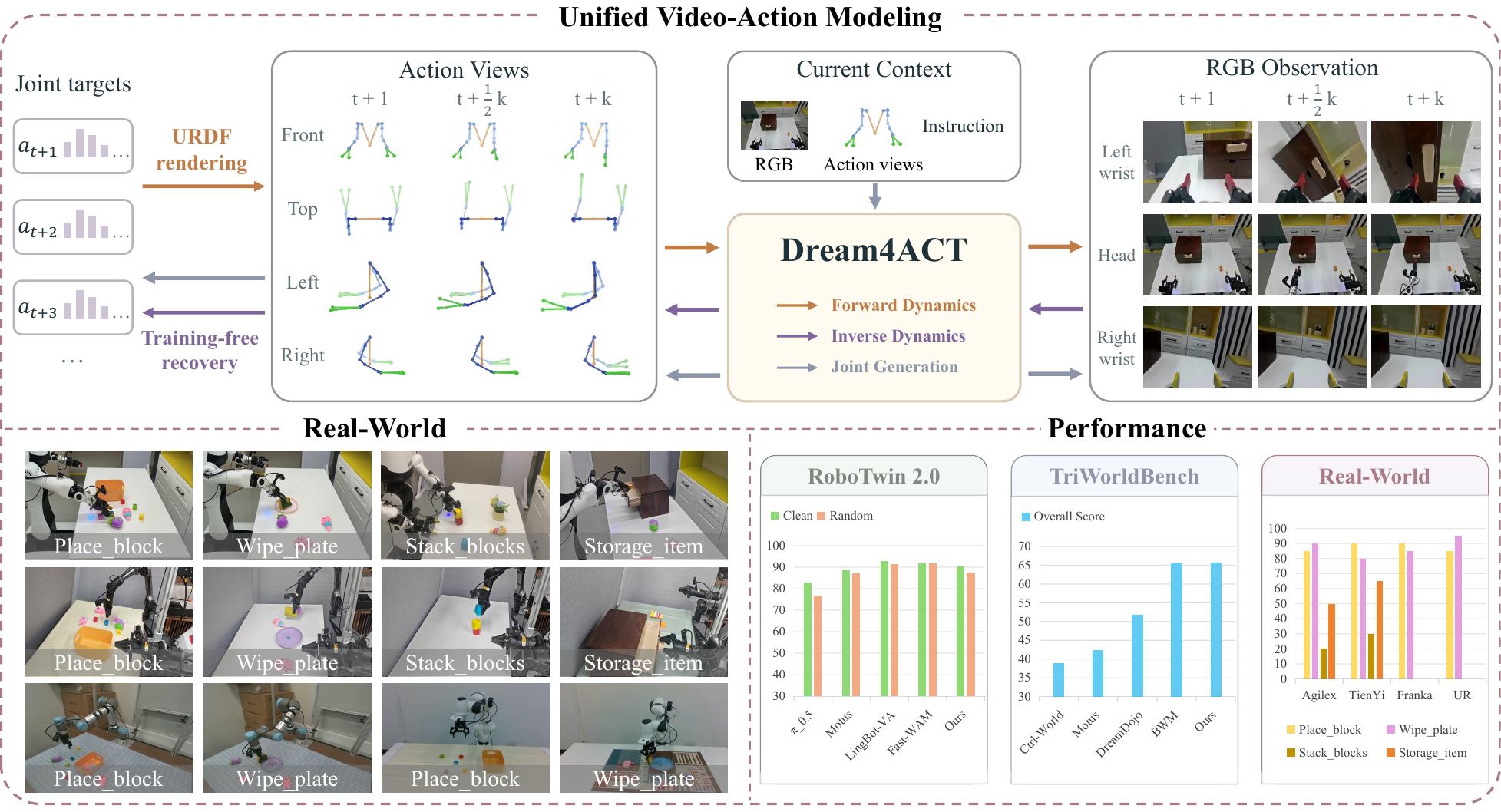}
    \caption{\textbf{Overview of Dream4ACT.} \textit{Top:} A shared action-view interface supports forward dynamics, inverse dynamics, and joint generation, with training-free recovery of joint targets. \textit{Bottom left:} Examples of multi embodiments real-world task execution. \textit{Bottom right:} Success rates and scores show strong performance in both robotic manipulation and action-conditioned video generation.}
    \vspace{-1.0em}
    \label{fig:teaser}
\end{figure*}

We instantiate this interface as \emph{action views}: multiview images of target joint configurations obtained through URDF-based forward kinematics and rendering. Unlike observation-aligned robot renderings~\cite{chen2026bridgev2w,gu2026geniworld}, our fixed virtual cameras are independent of physical observation cameras, removing the need for physical-camera extrinsic calibration in action-view construction. Saved camera presets remain fixed throughout training and inference, with consistent view roles across embodiments. A fixed tensor shape enables shared tokenization and prediction across embodiments, while the rendered articulated geometry preserves the information needed for configuration matching. Joint targets are then recovered by training-free, URDF-constrained optimization, without a learned embodiment-specific action decoder.

Building on this representation, \textbf{Dream4ACT} jointly models physical-camera observations and four action-view streams with a shared video autoencoder and diffusion transformer, conditioned on instruction-grounded scene features. Masked flow matching selects future conditioning and prediction streams, enabling forward dynamics, inverse dynamics, and joint generation with shared weights across jointly trained embodiments. Our model achieves 88.98\% average success on RoboTwin~2.0 and a TriWorldBench score of 65.66 using a separately trained checkpoint; closed-loop evaluations cover five simulated embodiments and four real-world platforms. A separately trained ablation compares action-view rendering with camera-aligned skeleton conditioning on the same recorded joint-state sequences. Together, these evaluations demonstrate the interface's strong capabilities both as a generated representation recoverable into executable commands and as a conditioning signal for future RGB prediction. 

In summary, our contributions are as follows:
\begin{itemize}[leftmargin=*,itemsep=2pt,topsep=2pt]
    \item We introduce \emph{action views}, a fixed-shape multiview representation of target joint configurations that allows heterogeneous joint spaces to share a common video-modeling interface while retaining embodiment-specific geometry.
    \item We develop a training-free, URDF-constrained multiview recovery to map predicted action views to joint targets without a learned embodiment-specific action decoder.
    \item Extensive experiments on simulation and real-world demonstrate that our model has strong performance on robotic manipulation and cross-view observation generation, with controlled ablation study validating the superior capability of generative visual quality.
\end{itemize}

\section{Related Work}
\label{sec:related_work}
\suppressfloats[t]

\paragraph{Video-Action World Models.}

Unified video-action models differ in how they couple modalities and select conditioning modes~\cite{wang2026world,wen2026dependency}. UVA~\cite{li2025unifiedvideoactionmodel} uses separate diffusion decoders with masked inputs, whereas UWM~\cite{zhu2025unified} and Pelican-Unify 1.0~\cite{zhang2026pelican} couple video and action diffusion with independent noise levels. The conditioning schedule and action representation are separate design choices: masking specifies which modalities are generated, whereas the representation determines how their contents encode control. Expert-based models coordinate modality-specific computation~\cite{bi2026motus,li2026causal}, while action-conditioned predictors such as Ctrl-World~\cite{guo2026ctrl} and RoboExpand~\citep{luo2026roboexpandreal2sim2realdataengine} generate multiview rollouts from numerical controls. Recent world action models further explore task generalization~\citep{shi-etal-2026-vla} and embodiment adaptation~\cite{ye2026world}, efficient inference~\cite{yuan2026fast}, and semantic or 3D alignment~\cite{li2026wall,yang20264d}. Joint prediction and flexible conditioning are thus prior capabilities, distinct from the choice of action representation.

\paragraph{Action Representations.}

Standardized vector interfaces support cross-embodiment policies~\cite{team2024octo,liu2025rdt}, whereas rendered representations additionally expose spatial robot geometry. Latent actions~\cite{bruce2024genie,wei2026causally} and visual-motion representations~\cite{ko2024learning,bi2026motus} offer alternatives to explicit joint vectors, but their conversion to executable commands depends on the chosen decoder or controller. For example, Hydra-0~\cite{li2026hydra} uses a trained action head, while Masked Visual Actions~\cite{alzayer2026masked} uses a learned inverse dynamics model. Visual interfaces differ in both the quantities they encode and their recovery mechanisms. Cosmos Policy~\cite{kim2026cosmos} encodes actions as latent frames without a separate action-generation architecture. Sharing the video backbone therefore does not necessarily imply an explicit geometric encoding of the robot. SpatialVAM~\cite{li2026spatialvamspatialawaremultiviewvideodiffusion} geometrically recovers end-effector positions from virtual-view heatmaps but learns rotation and gripper decoding. Action Images~\cite{zhen2026action} geometrically recovers end-effector pose and gripper commands from multiview images and supports joint generation, action-conditioned prediction, and action labeling. Separately, BridgeV2W~\cite{chen2026bridgev2w} and GeniWorld~\cite{gu2026geniworld} condition video prediction on URDF-rendered motion aligned with observation viewpoints, requiring the corresponding camera geometry. Dream4ACT instead renders full target joint configurations in separate action-view streams and approximately recovers joint targets through URDF-constrained multiview matching. Its prescribed virtual cameras remove physical-camera extrinsic calibration from action-view construction and recovery, while the URDF supplies embodiment-specific geometry without a learned per-robot decoder head.

\section{Method}
\label{sec:method}

\begin{figure*}[t]
    \centering
    \includegraphics[width=0.95\linewidth]{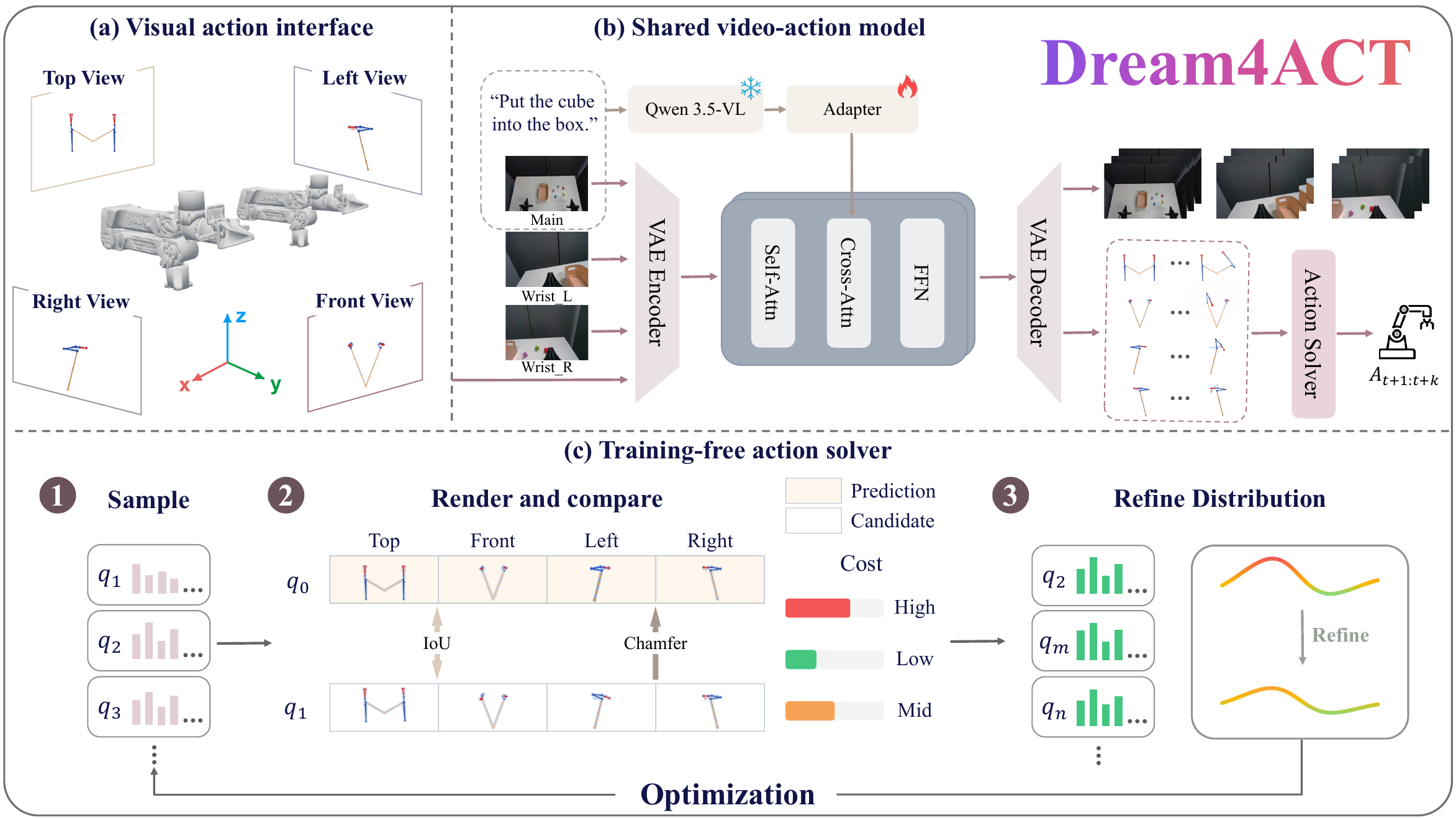}
    \caption{\textbf{Architecture of Dream4ACT.} \textit{(a)} URDF-based rendering produces four action views. \textit{(b)} A shared video VAE and diffusion transformer model RGB observations and action views with VLM-derived semantic conditioning, here we show the joint generation mode. \textit{(c)} Training-free multiview action solver recovers joint targets from predicted views.}
    \label{fig:overview}
\end{figure*}

\paragraph{Problem formulation.}
Our goal is to build a general video-action model. At time step $t$, the conditioning context is $c_t=(I_t,S_t,U,l)$, where $I_t$ comprises RGB images from head and wrist cameras, $S_t=(q_t,g_t)$ contains joint positions $q_t$ and gripper states $g_t$, $U$ is the URDF of the specific embodiment, and $l$ is the task instruction. Let $A_{t+1:t+k}$ denote the target joint configurations and gripper states over a prediction horizon $k$. We model $p_\theta(\cdot)$ in three modes, distinguished by which future sequences are observed or generated:
\begin{align}
&\text{Forward Dynamics:}\quad
p_\theta(I_{t+1:t+k}\mid c_t,A_{t+1:t+k}),\\
&\text{Inverse Dynamics:}\quad
p_\theta(A_{t+1:t+k}\mid c_t,I_{t+1:t+k}),\\
&\text{Joint Generation:}\quad
p_\theta(I_{t+1:t+k},A_{t+1:t+k}\mid c_t).
\end{align}
In practice, the action variables are represented by action views; predicted views are converted back to joint configurations through the recovery procedure.

\paragraph{Overview.}
Dream4ACT uses a shared visual action interface (Figure~\ref{fig:overview}). URDF rendering converts current and target joint configurations into four fixed virtual-camera action views (Section~\ref{sec:method:actionview}), while a trainable adapter over a frozen VLM extracts instruction-grounded conditions (Section~\ref{sec:method:vlm}). A shared video VAE separately encodes RGB and action-view sequences into latent tokens (Section~\ref{sec:method:tokens}). Conditioned on the adapter outputs, a diffusion transformer jointly models these tokens under masked flow matching, supporting forward dynamics, inverse dynamics, and joint generation (Section~\ref{sec:method:dit}). Finally, training-free URDF-constrained Gauss--Newton matching recovers joint targets from generated action views (Section~\ref{sec:method:recovery}).


\subsection{Visual Action Representation}
\label{sec:method:actionview}
We represent target states in image space to obtain a fixed-size action representation across embodiments with different dynamics and joint dimensions. Let $S_{t:t+k}$ denote the current and target configurations. For URDF $U$, action view $i$ at step $t'$ is
\begin{equation}
x_{t'}^i =
R\!\left(\operatorname{FK}(S_{t'};U);K_i,T_i\right),
\quad
i\in V,\quad t'=t,\ldots,t+k.
\label{eq:render}
\end{equation}

Here $R(\cdot)$ renders the articulated geometry, $\operatorname{FK}(\cdot)$ denotes forward kinematics and $K_i,T_i$ specify the intrinsics and extrinsics of virtual camera $i$. We use $|V|=4$ cameras whose coordinate frames and rendering conventions are shared across embodiments. The resulting action views have a fixed shape independent of joint dimensionality while retaining embodiment-specific geometry. They therefore provide heterogeneous joint spaces with a common visual interface for shared tokenization and video modeling. More details about action rendering are provided in Appendix~\ref{app:rendering}.


\subsection{Instruction-Grounded Semantic Adapter}
\label{sec:method:vlm}

We ground the task instruction $l$ in the current scene using a frozen pretrained vision-language model (VLM), Qwen3.5-VL~9B~\cite{bai2025qwen3}. It jointly processes $l$ and the current head-camera frame $I_t^{\mathrm{head}}$. For a VLM with $L$ layers, the last four hidden layers are partitioned into visual and language tokens:
\begin{equation}
h^{(\ell)}
=
\left[h_{\mathrm{vis}}^{(\ell)};
      h_{\mathrm{lang}}^{(\ell)}\right],
\quad\ell\in\mathcal{L}_{4}=\{L-3,L-2,L-1,L\}.
\end{equation}
Then we concatenate each token group along the feature dimension across layers and project it to the adapter dimension $d$:
\begin{equation}
H_{\mathrm{vis}}
=
P_{\mathrm{vis}}\!\left(
\operatorname{Concat}_{\ell\in\mathcal{L}_{4}}
h_{\mathrm{vis}}^{(\ell)}
\right),
\quad
H_{\mathrm{lang}}
=
P_{\mathrm{lang}}\!\left(
\operatorname{Concat}_{\ell\in\mathcal{L}_{4}}
h_{\mathrm{lang}}^{(\ell)}
\right),
\end{equation}
$P_{\mathrm{vis}}(\cdot)$ and $P_{\mathrm{lang}}(\cdot)$ denote visual and language projection layers separately. Then the adapter uses four learnable queries $Q\in\mathbb{R}^{4\times d}$ to compress the visual features:
\begin{equation}
\bar{Q}
=
\operatorname{CrossAttn}
\left(Q,H_{\mathrm{vis}},H_{\mathrm{vis}}\right).
\end{equation}
The result is concatenated with the language tokens along the sequence dimension and refined by joint self-attention:
\begin{equation}
F_{\mathrm{sem}}
=
\operatorname{SelfAttn}
\left([\bar{Q};H_{\mathrm{lang}}]\right)
\in\mathbb{R}^{N_c\times d},
\quad
N_c=4+N_{\mathrm{lang}}.
\label{eq:condition}
\end{equation}
Here, $N_{\mathrm{lang}}$ is the language-token count. Adapter's structure is presented in Figure~\ref{fig:adapter}
, it is trained jointly with the generative backbone.


\subsection{Multi-Sequence Latent Tokenization}
\label{sec:method:tokens}

Over the window $t:t+k$, the model processes multi physical RGB observations $I_{t:t+k}$ and four action views $x_{t:t+k}$. We index the multi sequences by $s\in\mathcal{S}=\mathcal{S}_{\mathrm{rgb}}\mathbin{\cup}V$ and denote each by $y^s_{t:t+k}$, where $y^s=I^s$ for $s\in\mathcal{S}_{\mathrm{rgb}}$ and $y^s=x^s$ for $s\in V$. A shared Wan2.2 VAE encoder~\cite{wan2025wan} $\mathcal{E}$ is applied to each sequence separately,
\begin{equation}
z_{s} = \mathcal{E}(y^{s}_{t:t+k}),
\quad s \in \mathcal{S},
\label{eq:vae}
\end{equation}
The latent grid is indexed by temporal positions $f$ and spatial positions $p$. Here $f=0$ is the current frame latent and $f\geq1$ denotes future latents. Before joint processing, each latent feature receives a modality and view embedding,
\begin{equation}
\tilde{z}_{s} = z_{s} + e^{\mathrm{mod}}_{\mu(s)} + e^{\mathrm{view}}_{s},
\quad s \in \mathcal{S},
\label{eq:annotate}
\end{equation}
where $\mu(s)\in\{\mathrm{rgb},\mathrm{action}\}$ selects one of two modality embeddings and $e^{\mathrm{view}}_s$ identifies one of multi sequence slots. These embeddings are shared across embodiments. For rotary position embeddings (RoPE)~\cite{su2024roformer}, the token at latent frame $f$ and spatial position $p$ receives
\begin{equation}
\phi(s, f, p) = (f,\; p + \delta_{s}),
\label{eq:canvas}
\end{equation}
where $\delta_s$ places sequence $s$ in a distinct spatial region. This shares the temporal coordinate while preventing spatial-position collisions across sequences. The annotated grids are then flattened and concatenated as $z=[\tilde z_s]_{s\in\mathcal S}$ to build the input of diffusion transformer.


\subsection{Joint Diffusion Transformer}
\label{sec:method:dit}
We choose a diffusion transformer (DiT)~\cite{peebles2023scalable} as our generative backbone. Each block applies joint self-attention across all input sequences, then makes cross-attention to the semantic condition $F_{\mathrm{sem}}$ in Equation~\ref{eq:condition}, and finally a position-wise feed-forward network (FFN).

Let $L$ denote the sequence length of $s$, we extend conditional flow matching~\cite{lipman2022flow,liu2022flow} with modality-specific noise levels and a binary mask $m\in\{0,1\}^{|\mathcal S|\times L}$. Sequence $s$ uses $\tau_s=\tau_{\mathrm{rgb}}$ for $s\in\mathcal S_{\mathrm{rgb}}$ and $\tau_s=\tau_{\mathrm{act}}$ for $s\in V$. Clean and corrupted latents are represented jointly as
\begin{equation}
\tilde z^{\tau,m}_{s,f}
=
(1-m_{s,f})z_{s,f}
+m_{s,f}\big[\tau_s z_{s,f}+(1-\tau_s)\epsilon_{s,f}\big],
\quad\epsilon_{s,f}\sim\mathcal N(0,\mathbf I),
\label{eq:interpolant}
\end{equation}
where $m_{s,f}=0$ preserves the data and $m_{s,f}=1$ follows the linear path from noise at $\tau_s=0$ to data at $\tau_s=1$. The velocity field $v_\theta$ is trained only on corrupted positions:
\begin{equation}
\mathcal L
= \mathbb E_{z,\epsilon,\boldsymbol\tau,m}
\Biggl[
    \sum_{s\in\mathcal S}\sum_{f=0}^{L-1}
    m_{s,f}
    \Bigl\|
        \bigl[
            v_\theta\!\bigl(
                \tilde z^{\boldsymbol\tau,m};
                \boldsymbol\tau,m,F_{\mathrm{sem}}
            \bigr)
        \bigr]_{s,f}
        - \bigl(z_{s,f}-\epsilon_{s,f}\bigr)
    \Bigr\|_2^2
\Biggr].
\label{eq:loss}
\end{equation}
where $\boldsymbol\tau=(\tau_s)_{s\in\mathcal S}$ collects the sequence-wise noise levels, $[\cdot]_{s,f}$ selects the output for sequence $s$ at latent frame $f$. The mask also selects the operating mode while keeping the current RGB and action-view latents at $f=0$ clean:
\begin{equation}
m_{s,f}=\mathbbm{1} [s\in M\wedge f\geq1],
\quad M\subseteq\mathcal S.
\label{eq:mask}
\end{equation}
One of the three choices of $M$ in Table~\ref{tab:schedule} is sampled for each training example.


\subsection{Training-Free Action Recovery}
\label{sec:method:recovery}

We convert action views predicted by IDM and JGM into joint targets by fitting URDF-derived geometry to the predictions across four virtual cameras. All reported recovery and robot-execution results use the solver in Table~\ref{tab:recovery}, without a learned embodiment-specific decoder.

At local frame $j=0,\ldots,k$, let $u_j=(q_j,g_j)$ contain joint positions and provisional gripper states, with $u_0=S_t$. For each candidate, forward kinematics determines the robot geometry, which we project using the same virtual cameras as in action-view construction. Image fields extracted from the predictions provide the matching evidence. Let $r_j(u_j)$ collect the residuals between the projected candidate geometry and these fields across the four views. With residual weights $W_j$, we minimize
\begin{equation}
E_j^{\mathrm{field}}(u_j)
=\tfrac{1}{2}r_j(u_j)^\top W_jr_j(u_j)
\label{eq:recovery-field-cost}
\end{equation}
by refining multiple candidate initializations in parallel with batched Levenberg-Marquardt (LM). With $J_j=\partial r_j/\partial u_j$, each update solves
\begin{equation}
\left(J_j^\top W_jJ_j+\mu M\right)\Delta u_j=-J_j^\top W_jr_j,
\label{eq:recovery-lm}
\end{equation}
where $\mu>0$ controls damping and $M$ is a positive-definite damping metric.

The frame-wise estimates are refined by alternating regularized cubic B-spline trajectory fitting with finer-scale LM updates. After each fit, we enforce joint limits and restore $u_0=S_t$. These recovered configurations are then assembled into the executable action chunk  $\hat{A}_{t+1:t+k}$. Finally, we render the action sequence, compute the IoU and the Chamfer metric for evaluation. Appendix~\ref{app:recovery} provides more implementation details.

\begin{table}[t]
    \caption{Algorithm for training-free action recovery.}
    \label{tab:recovery}
    \centering
    \small
    \begin{tabular}{@{}p{0.96\linewidth}@{}}
    \toprule
    \textbf{Input:} Action views $\hat{x}=\{\hat{x}_{t+j}^{i}\}_{i\in V,\,j=1:k}$;
    URDF $U$; camera configuration $\mathcal C=\{K_i,T_i\}_{i\in V}$; state $S_t$. \\
    \textbf{Output:} Recovered action chunk $\hat{A}_{t+1:t+k}$. \\
    \midrule
    $\Phi\gets\operatorname{ImageFields}(\hat{x}),\quad \hat{A}_t\gets S_t$. \\
    $\hat{A}_{t+1:t+k}\gets\operatorname{BatchedLM}(\Phi;U,\mathcal C,S_t)$. \\
    $\hat{A}_{t:t+k}\gets\operatorname{RefineChunk}(\hat{A}_{t:t+k};\hat{x},U,\mathcal C)$.
    \hfill\emph{B-spline} \\
    $\hat{A}_{t+j}\in\mathcal Q(U),\quad \hat{A}_t=S_t$.
    \hfill\emph{refinement constraints} \\
    $E\gets\operatorname{Score}\!\left(R(\hat{A}_{t+1:t+k};U,\mathcal C),\hat{x}\right)$.
    \hfill\emph{evaluation} \\
    \textbf{return} $\hat{A}_{t+1:t+k}$. \\
    \bottomrule
    \end{tabular}
\end{table}

\section{Experiments}
\label{sec:experiments}

We evaluate our model along three complementary axes: closed-loop manipulation, multi-embodiment capability and action-conditioned multiview prediction. 

\subsection{Simulation}
\label{sec:experiments:simulation}

\paragraph{Benchmarks.}
We evaluate closed-loop manipulation capability on RoboTwin~2.0~\cite{chen2025robotwin}, a simulation benchmark comprising 50 bimanual manipulation tasks. We collect 2,500 clean and 25,000 randomized demonstrations, corresponding to 50 and 500 demonstrations per task, respectively. Randomization varies backgrounds, tabletop objects, and lighting.

To evaluate action-conditioned world modeling, we use TriWorldBench~\cite{liu2026triworldbench}, which assesses generated videos from synchronized head and wrist cameras. We use the checkpoint fine-tuned on the benchmark and run inference in forward dynamics mode. Given the evaluation action trajectories, we render the corresponding action-view sequences for conditioning.  Following the official submission protocol, we generate corresponding head and wrist videos per episode, with frame counts matching the supplied trajectory lengths, for official evaluation. 

\paragraph{Baselines.}
For manipulation, we compare against $\pi_{0.5}$~\cite{intelligence2025pi_}, Motus~\cite{bi2026motus}, LingBot-VA~\cite{li2026causal}, and Fast-WAM~\cite{yuan2026fast}, covering vision-language-action policies, unified video-action models, and world action models. We report publicly available success rates for these methods. For action-conditioned visual prediction on TriWorldBench, we include Ctrl-World~\cite{guo2026ctrl}, Motus~\cite{bi2026motus}, Genie Envisioner~\cite{liao2026genie}, DreamDojo~\cite{gao2026dreamdojo}, and BWM~\cite{bwm2026bwm} as strong baselines.

\paragraph{Metrics.}
For manipulation, we follow RoboTwin~2.0 official task-success criteria and evaluate each of the 50 tasks over 100 trials per configuration. For TriWorldBench, we report its six aggregate dimensions: tri-view consistency (TVC), task alignment (TA), physical and 3D coherence (P3D), motion quality (MQ), temporal consistency (TC), and visual quality (VQ), together with the official overall TWB-Score. Higher scores are better for all these reported metrics. TVC measures agreement across observation views, whereas TA assesses task alignment; these video scores complement the executable-task success rates.

\begin{table}[t]
    \centering
    \small
    \setlength{\tabcolsep}{10pt}
    \caption{Success rates (\%) on RoboTwin2.0 over 50 clean and 50 randomized tasks. Table~\ref{tab:robotwin_full} 
    presents a comparison of success rates for each individual task.}
    \begin{tabular}{lccc}
        \toprule
        \textbf{Method} & Clean & Randomized & \textbf{Avg.} \\
        \midrule
        $\pi_{0.5}$~\cite{intelligence2025pi_}     
        & 82.74      & 76.76    & 79.80   \\
        Motus~\cite{bi2026motus}           
        & 88.66      & 87.02    & 87.84   \\
        LingBot-VA~\cite{li2026causal}   
        & 92.90      & 91.50    & \textbf{92.20}  \\
        Fast-WAM~\cite{yuan2026fast}      
        & 91.88      & 91.78    & 91.83   \\
        \textbf{Ours}   
        & 90.50      & 87.46    & 88.98  \\
        \bottomrule
    \end{tabular}
    \label{tab:robotwin_results}
\end{table}

\begin{table*}[t]
\centering
\small
\caption{TriWorldBench results. Dream4ACT is evaluated on the official 500-episode test set and reports six key indicators.}
\label{tab:triworldbench}
    \begin{tabular}{lccccccc}
        \toprule
        Model & TVC & TA & P3D & MQ & TC & VQ & Overall \\
        \midrule
        Ctrl-World~\cite{guo2026ctrl}
        & $57.42$ & $43.72$ & $34.67$ 
        & $29.28$ & $46.17$ & $16.64$ & $38.98$ \\
        Motus~\cite{bi2026motus}
        & $66.70$ & $49.69$ & $34.60$ 
        & $24.63$ & $26.56$ & $16.26$ & $42.35$ \\
        Genie Envisioner~\cite{liao2026genie}
        & $62.39$ & $33.17$ & $54.00$ 
        & $20.17$ & $32.18$ & $17.46$ & $40.73$ \\
        DreamDojo~\cite{gao2026dreamdojo}
        & $69.63$ & $56.24$ & $43.84$ 
        & $27.96$ & $60.84$ & $21.02$ & $51.72$ \\
        BWM~\cite{bwm2026bwm}
        & $\textbf{81.87}$ & $\mathbf{86.05}$ & $60.40$ 
        & $41.29$ & $62.81$ & $\textbf{31.42}$ & $65.54$ \\
        \midrule
        \textbf{Ours}
        & $81.63$ & $84.22$ & $\mathbf{61.30}$ 
        & $\textbf{41.66}$ & $\textbf{64.88}$ & $\textbf{31.42}$ & $\textbf{65.66}$ \\
        \bottomrule
    \end{tabular}
\end{table*}

\paragraph{Results.}
For closed-loop control in simulation, we use joint generation in image-to-video (I2V) mode with 8 denoising steps, predicting both RGB videos and action views. On RoboTwin 2.0, we use 40-step action chunks. Model inference and action recovery take 3.8 s and 1.6 s, respectively, totaling approximately 5.4 s per chunk.

Table~\ref{tab:robotwin_results} shows that our model achieves success rates of 90.50\% and 87.46\% under clean and randomized settings, respectively, averaging 88.98\%. Compared with publicly reported results, Dream4ACT exceeds $\pi_{0.5}$ by 7.76 and 10.7 percentage points and is slightly higher than Motus, while LingBot-VA and Fast-WAM achieve higher scores. These results demonstrate effective closed-loop manipulation through generated action views and training-free recovery. The full 50-task evaluation uses a separately trained checkpoint. The multi-embodiment evaluation uses a checkpoint jointly trained across five embodiments, with its parameters fixed across all five robots.

In forward dynamics mode, Dream4ACT achieves a TWB-Score of 65.66, comparable to BWM's 65.54 (Table~\ref{tab:triworldbench}). Among the listed methods, our method obtains the highest reported P3D, MQ, and TC scores and ties BWM on VQ. Its TVC and TA scores are 81.63 and 84.22, respectively, compared with BWM's 81.87 and 86.05, showing competitive cross-view consistency and task alignment. Together with the manipulation results, they support the dual role of action views as conditioning inputs for multiview prediction and as generated representations recoverable into executable commands.

\begin{table*}[t]
    \centering
    \small
    \setlength{\tabcolsep}{3pt}
    \caption{
        Success rates (\%) of the same jointly trained checkpoint across five embodiments on 31 shared RoboTwin~2.0 tasks.
    }
    \label{tab:robotwin_multi_embodiment}
    \begin{tabular}{l *{4}{c@{\hspace{3pt}}c@{\hspace{3pt}}c}}
        \toprule
        \multirow{2}{*}{\textbf{Method}}
          & \multicolumn{3}{c}{ \textbf{Octo}~\cite{team2024octo}}
          & \multicolumn{3}{c}{ \textbf{RDT}~\cite{liu2025rdt}}
          & \multicolumn{3}{c}{ \textbf{Ours w/ joint action}}
          & \multicolumn{3}{c}{\textbf{Ours}}  \\
        \cmidrule(lr){2-4}
        \cmidrule(lr){5-7}
        \cmidrule(lr){8-10}
        \cmidrule(lr){11-13}
          &  Clean   &  Random   &  Avg.
          &  Clean   &  Random   &  Avg.
          &  Clean   &  Random   &  Avg.
          & Clean  & Random  & Avg. \\
        \midrule
        Aloha-Agilex  
        & 18.32 &  18.88  &  18.60
        &  10.97  &  8.90   &  9.94
        &  35.61  &  34.71  &  35.16
        & \textbf{81.94}  & \textbf{82.45}  & \textbf{82.19}  \\ 
        Piper
        &  17.29  &  21.29  &  19.29
        &  10.27  &  11.87  &  11.07
        &  38.84  &  37.68  &  38.26
        & \textbf{81.55}  & \textbf{82.32}  & \textbf{81.94}  \\
        ARX-X5
        &  17.03  &  18.45  &  17.74
        &  13.03  &  12.65  &  12.84
        &  38.06  &  36.52  &  37.29
        & \textbf{82.19}  & \textbf{80.26}  & \textbf{81.23}  \\
        Franka-Panda
        &  16.13  &  17.94  &  17.04
        &  10.58  &  12.00  &  11.29
        &  26.06  &  22.58  &  24.32
        & \textbf{64.52}  & \textbf{62.45}  & \textbf{63.48}  \\
        UR5-Xsg
        &  15.48 &  17.29 &  16.39
        &  13.55  &  16.65  &  15.10
        &  \textbf{36.13}  &  \textbf{31.35}  &  \textbf{33.74}
        & 31.87 & 30.06 & 30.97 \\
        \bottomrule
    \end{tabular}
\end{table*}

\begin{table*}[h]
    \centering
    \small
    \caption{Real-world success rates (\%) using the same jointly trained checkpoint across four robot platforms. Each embodiment--task pair is evaluated over 20 trials under scene variations, including randomized background and randomized distractor objects. \textbf{Common} averages the first two tasks; \textbf{All} averages all tasks evaluated on each platform. ``--'' denotes a task not evaluated on that platform.}
    \begin{tabular}{lcccccc}
        \toprule
        \textbf{Robot} & Place\_block & Wipe\_plate & Stack\_blocks & Storage\_item & \textbf{Common} & \textbf{All} \\
        \midrule
        Aloha-Agilex    & 85.0    & 90.0    & 20.0   
                        & 50.0    & 87.5    & 61.3 \\
        TienYi2.5 Pro   & 90.0    & 80.0    & 30.0   
                        & 65.0    & 85.0    & 66.3 \\
        Franka          & 90.0    & 85.0    & --    
                        & --      & 87.5    & -- \\
        UR5e            & 85.0    & 95.0    & --    
                        & --      & 90.0    & -- \\
        \bottomrule
    \end{tabular}
    \label{tab:real_world_experiments}
\end{table*}

\paragraph{Multi-embodiment evaluation.}
\label{sec:experiments:sim_cross_embodiment}
We evaluate the jointly trained checkpoint across five embodiments: Aloha-Agilex, ARX-X5, Franka-Panda, Piper, and UR5-Xsg. To keep task composition consistent, we use the intersection of tasks with available demonstrations for all five robots, yielding 31 tasks from the official 50-task suite. This subset defines the multi-embodiment training component and its evaluation task set, separate from the full 50-task evaluation above. During training, we use 50 clean and 500 randomized episodes for each embodiment-task pair. For test, we run 25 trials per task per robot under each clean and randomized setting, totaling 775 trials per setting, and average success rates equally across tasks. Model parameters remain unchanged across all five embodiments, with the corresponding URDF used for action rendering and recovery.

As shown in Table~\ref{tab:robotwin_multi_embodiment}, we compare with Octo~\cite{team2024octo} and RDT~\cite{liu2025rdt} as unified action-space baselines, each is trained for 100k steps with a batch size of 16. Our method substantially outperforms the baseline, achieving average success rates above 81\% on Aloha-Agilex, ARX-X5, and Piper, and 63.48\% and 30.97\% on Franka-Panda and UR5-Xsg, respectively. These results demonstrate that a single jointly trained checkpoint supports executable control across distinct kinematic structures through a common visual action interface. More analysis about recovery error is provided in Appendix~\ref{app:multi_embodiments}.

\begin{wraptable}{r}{0.40\linewidth}
    \centering
    \small
    \vspace{-1.0em}
    \caption{Analysis of the robustness of OOD tasks execution capability.}
    \begin{tabular}{lccc}
        \toprule
        \multirow{2}{*}{\textbf{Method}}
          & \multicolumn{3}{c}{\textbf{RoboTwin~2.0}} \\
          \cmidrule(lr){2-4}
          & Clean
          & Random
          & Avg. \\
        \midrule
        Ours           
        & 19.37 & 17.47 & 18.42  \\
        \bottomrule
    \end{tabular}
    \label{tab:ood_test}
\end{wraptable}

\paragraph{OOD evaluation.}
To assess the robustness of camera rig and out-of-distribution capability, we construct camera intrinsics using a fixed field of view (FOV), without relying on task-specific parameters, and evaluate the checkpoint trained on 31 tasks across five robot embodiments on the remaining 19 Aloha tasks through closed-loop execution. For each task, we conduct 25 trials under each of the clean and randomized settings and report the corresponding success rates in Table~\ref{tab:ood_test}.

\subsection{Real-World Experiments}
\label{sec:experiments:real_world}

\paragraph{Settings.}
We evaluate a single jointly trained checkpoint on two bimanual platforms, Aloha-Agilex and TienYi2.5 Pro, and two single-arm platforms, Franka Research 3 and UR5e. All four platforms are evaluated on \texttt{place\_block} and \texttt{wipe\_plate}; the bimanual platforms are additionally evaluated on \texttt{stack\_blocks} and \texttt{storage\_item}. We collect 60 demonstrations per embodiment--task pair with randomized object placements, totaling 720 demonstrations across 12 pairs. Demonstrations use the nominal background for each setup, without background variation or additional randomized distractor objects. Bimanual platforms use one head camera and two wrist cameras, whereas single-arm platforms use one wrist camera and one external camera. Model parameters remain unchanged across platforms, with each robot's URDF used for rendering and action recovery.

\begin{figure}[t]
    \centering
    \includegraphics[width=1.0\linewidth]{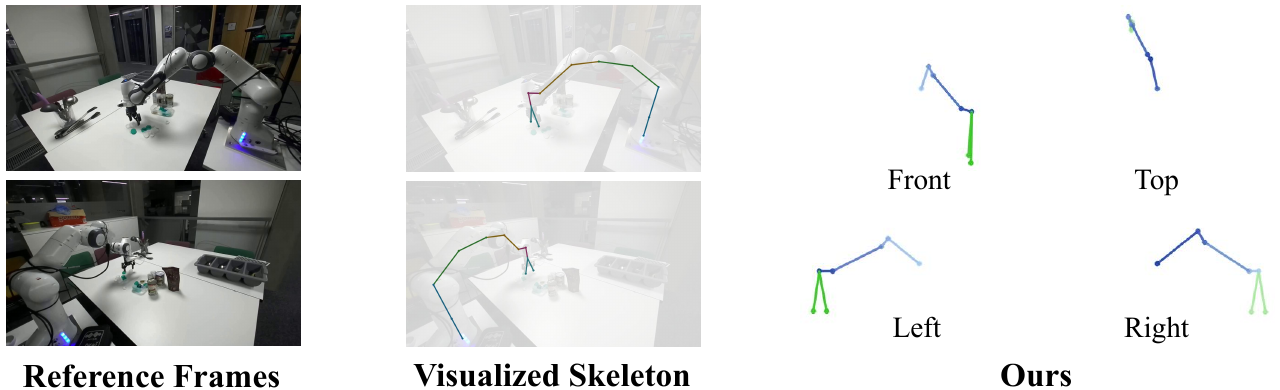}
    \caption{\textbf{Comparison of conditioning representations.} The same recorded joint configurations are represented as camera-aligned skeleton projections or action-view renderings from four prescribed virtual cameras.}
    \vspace{-0.5em}
    \label{fig:ablation}
\end{figure}

\paragraph{Results.}
On the two common tasks, Aloha-Agilex, TienYi2.5 Pro, Franka Research 3, and UR5e achieve mean success rates of 87.5\%, 85.0\%, 87.5\%, and 90.0\%, respectively (Table~\ref{tab:real_world_experiments}). The extended bimanual evaluations additionally cover stacking and storage, yielding four-task averages of 61.3\% and 66.3\% for Aloha-Agilex and TienYi2.5 Pro. These results demonstrate that the same checkpoint
supports real-world manipulation across single-arm and bimanual platforms with different observation configurations, using a common visual action interface and training-free recovery without learned embodiment-specific action heads. More details are provided in Appendix~\ref{app:real-world}.

\begin{table*}[t]
    \centering
    \small
    \caption{ RGB prediction fidelity and Subject consistency on 50 held-out DROID trajectories from CLVR and RAD. Both variants condition on recorded joint states and predict 41-frame sequences.}
    \label{tab:ablation_study}
    \begin{tabular}{lcccc}
        \toprule
        \textcolor{black}{\textbf{Method}}
        & \textcolor{black}{\textbf{PSNR}$\uparrow$}
        & \textcolor{black}{\textbf{SSIM}$\uparrow$}
        & \textcolor{black}{\textbf{LPIPS}$\downarrow$}
        &   \textbf{Subject Consistency}$\uparrow$ \\
        \midrule
        \textcolor{black}{Skeleton rendering}
        & \textcolor{black}{23.190} & \textcolor{black}{0.899} & \textcolor{black}{0.105} &   0.0885 \\
        \midrule
          One action-view
        &   23.621 &   0.880 &   0.114 &   0.1083 \\
          Two action-views
        &   24.707 &   0.890 &   0.106 &   0.1068 \\
          Three action-views
        &   24.671 &   0.889 &   0.106 &   0.1066 \\
          \textbf{Four action-views (Ours)}
        &   \textbf{24.780} &   \textbf{0.900} &   \textbf{0.102} &   \textbf{0.1239} \\
        \bottomrule
    \end{tabular}
\end{table*}

\subsection{Ablation Study}
\label{sec:experiments:ablation}

\paragraph{Action representation ablation.}
We compare camera-aligned skeleton conditioning with our fixed-view robot rendering for RGB prediction on DROID~\cite{khazatsky2024droid} dataset. Both representations are constructed from the same recorded joint-state sequences. Following the camera-refinement procedure used in PointWorld~\cite{huang2026pointworld}, the baseline projects the Franka skeleton into the observation views using refined physical-camera calibration. Our representation instead renders the articulated robot from four prescribed virtual cameras, without requiring physical-camera extrinsics for its construction. Figure~\ref{fig:ablation} provides a visual comparison.

We train two separate variants on the same 1,000 trajectories uniformly sampled from DROID sources excluding CLVR and RAD. Both variants are evaluated using the 10k checkpoint. These ablation models are trained from scratch. Evaluation uses 50 trajectories uniformly sampled from the held-out CLVR/RAD pool. Both variants predict 41-frame RGB sequences conditioned on recorded joint states. We compute PSNR, SSIM, LPIPS and Subject Consistency~\cite{liu2026triworldbench} (measures whether robot and object identity and appearance remain stable across frames) against temporally aligned ground-truth physical-camera observations and average the scores over the test set.

Both variants in Table~\ref{tab:ablation_study} use the same two third-person RGB observations. The skeleton baseline uses two camera-aligned skeleton views, whereas our model uses a variable number of action views, with the view count sampled uniformly during training. Skeleton views use the same pipeline as action views, with view embeddings identifying viewpoints and the action modality embedding identifying skeleton tokens. During training and inference, we use \(320\times224\) resolution as input and \(640\times480\) resolution as output. We also ablate the number of action views for forward dynamics mode. We find it is interesting that training with mixed action-views is helpful for video generation.

We compare skeleton rendering with variants using one to four action-views, added in the order of Front, Left, Top and Right. As shown in Table~\ref{tab:ablation_study}, visual prediction quality generally improves as the number of action views increases. Four action-views (Ours) outperforms skeleton rendering across all four metrics and achieves the best results among the evaluated variants. The improvements across all four metrics support action-view rendering as an effective conditioning representation for RGB prediction, without requiring alignment of the conditioning views to the physical cameras.

\begin{table*}[t]
    \centering
    \small
    \caption{TriWorldBench ablation study. GT-dependent metrics are excluded from TVC, TA and MQ. Overall is the equal-weight mean of the 11 shared metrics.}
    \label{tab:triworldbench_representation}
    \begin{tabular}{lccccccc}
        \toprule
        Model & TVC & TA & P3D & MQ & TC  & Overall \\
        \midrule
         Ours w/ numerical action conditioning
        &  79.76
        &  54.23
        &  51.59
        &  41.54
        &  59.20
        &  59.76 \\
        \textbf{Ours}
        & \textbf{89.54}
        & \textbf{74.65}
        & \textbf{61.30}
        & \textbf{42.42}
        & \textbf{64.88}
        & \textbf{67.76} \\
        \bottomrule
    \end{tabular}
\end{table*}

\paragraph{Padded joint-vector ablation.}
We further ablate our action representation on RoboTwin~2.0 using a unified 16-dimensional joint vector: seven joint positions and one gripper state per arm, with dimensions 7 and 15 masked for 6-DoF arms. Following RDT~\cite{liu2025rdt}, we use an MLP to encode each timestep’s action into one token, and then concatenated with RGB video tokens, another MLP predicts the full joint vector. We train this variant for 100k steps on 31 tasks shared across 5 embodiments and evaluate 25 trials per embodiment-task pair in each of the clean and randomized settings. As shown in Table~\ref{tab:robotwin_multi_embodiment}, our method is higher than the variant on Aloha-Agilex, Piper, ARX-X5 and Franka-Panda, while slightly lower on UR5-Xsg.

\paragraph{TriWorldBench ablation.}
Since GT-dependent metrics require submission to the official evaluation system, we report only the GT-independent metrics evaluated locally: TVC (tri-view consistency), TA (task alignment), P3D (physical and 3D coherence), MQ (motion quality), and TC (temporal consistency). The Overall score is a weighted aggregate of these reported metrics. As shown in Table~\ref{tab:triworldbench_representation}, our method outperforms numerical action conditioning on every reported metric.

\section{Conclusion}
\label{sec:conclusion}
We presented Dream4ACT, a video--action world model for joint-controlled manipulation that represents target joint configurations as fixed-shape action views. Masked flow matching supports forward dynamics, inverse dynamics, and joint generation, while training-free URDF-constrained matching recovers joint targets for execution. Experiments on RoboTwin2.0 and real robots show that the same weights support closed-loop control across multiple jointly trained embodiments and our forward dynamics provides high quality task-aligned multi-view generation.

\newpage
\bibliography{reference.bib}

@article{wan2025wan,
  title={Wan: Open and advanced large-scale video generative models},
  author={Wan, Team and Wang, Ang and Ai, Baole and Wen, Bin and Mao, Chaojie and Xie, Chen-Wei and Chen, Di and Yu, Feiwu and Zhao, Haiming and Yang, Jianxiao and others},
  journal={arXiv preprint arXiv:2503.20314},
  year={2025}
}

@article{chen2025robotwin,
  title={Robotwin 2.0: A scalable data generator and benchmark with strong domain randomization for robust bimanual robotic manipulation},
  author={Chen, Tianxing and Chen, Zanxin and Chen, Baijun and Cai, Zijian and Liu, Yibin and Li, Zixuan and Liang, Qiwei and Lin, Xianliang and Ge, Yiheng and Gu, Zhenyu and others},
  journal={arXiv preprint arXiv:2506.18088},
  year={2025}
}

@article{li2026causal,
  title={Causal world modeling for robot control},
  author={Li, Lin and Zhang, Qihang and Luo, Yiming and Yang, Shuai and Wang, Ruilin and Han, Fei and Yu, Mingrui and Gao, Zelin and Xue, Nan and Zhu, Xing and others},
  journal={arXiv preprint arXiv:2601.21998},
  year={2026}
}

@article{intelligence2025pi_,
  title={$\pi_{0.5}$: a Vision-Language-Action Model with Open-World Generalization},
  author={Intelligence, Physical and Black, Kevin and Brown, Noah and Darpinian, James and Dhabalia, Karan and Driess, Danny and Esmail, Adnan and Equi, Michael and Finn, Chelsea and Fusai, Niccolo and others},
  journal={arXiv preprint arXiv:2504.16054},
  year={2025}
}

@article{yuan2026fast,
  title={Fast-wam: Do world action models need test-time future imagination?},
  author={Yuan, Tianyuan and Dong, Zibin and Liu, Yicheng and Zhao, Hang},
  journal={arXiv preprint arXiv:2603.16666},
  year={2026}
}

@inproceedings{bi2026motus,
  title={Motus: A unified latent action world model},
  author={Bi, Hongzhe and Tan, Hengkai and Xie, Shenghao and Wang, Zeyuan and Huang, Shuhe and Liu, Haitian and Zhao, Ruowen and Feng, Yao and Xiang, Chendong and Rong, Yinze and others},
  booktitle={Proceedings of the IEEE/CVF Conference on Computer Vision and Pattern Recognition},
  pages={35101--35113},
  year={2026}
}

@article{alzayer2026masked,
  title={Masked Visual Actions for Unified World Modeling},
  author={Alzayer, Hadi and Huang, Wenlong and Chen, Haonan and Luey, Christopher and Zhang, Lvmin and Agrawala, Maneesh and Wetzstein, Gordon and Fei-Fei, Li and Du, Yilun and Wu, Jiajun and others},
  journal={arXiv preprint arXiv:2607.19343},
  year={2026}
}

@inproceedings{peebles2023scalable,
  title={Scalable diffusion models with transformers},
  author={Peebles, William and Xie, Saining},
  booktitle={2023 IEEE/CVF International Conference on Computer Vision (ICCV)},
  pages={4172--4182},
  year={2023},
  organization={IEEE}
}

@article{lipman2022flow,
  title={Flow matching for generative modeling},
  author={Lipman, Yaron and Chen, Ricky TQ and Ben-Hamu, Heli and Nickel, Maximilian and Le, Matt},
  journal={arXiv preprint arXiv:2210.02747},
  year={2022}
}

@article{bwm2026bwm,
  title={BWM: A Low-Cost High-Fidelity World Simulator for Robot Learning},
  author={BWM Team and others},
  journal={arXiv preprint arXiv:2607.29302},
  year={2026}
}

@article{li2026hydra,
  title={Hydra-0: Action Flow for Generalist World Modeling and Control},
  author={Li, Hongyu and Wen, Bowen and Zhu, Xinghao and Wang, Yixuan and Du, Yilun and Li, Yunzhu and Konidaris, George and Birchfield, Stan and Pouya, Soha and Li, Chenran and others},
  journal={arXiv preprint arXiv:2608.18077},
  year={2026}
}

@article{wei2026causally,
  title={Causally Debiased Latent Action Model for Embodied Action Conditioned World Models},
  author={Wei, Yufan and Zhou, Kun and Mao, Lingjun and Zhang, Zijun and Xu, Ziming and Xi, Ziqiao and Liang, Shuang and Han, Ruobing and Yan, Yuchen and Wang, Xinyue and others},
  journal={arXiv preprint arXiv:2607.09185},
  year={2026}
}

@article{gu2026geniworld,
  title={GeniWorld: A Generalizable Interactive World Model for Robotic Manipulation via Visual Actions},
  author={Gu, Chenghao and Yu, Hanyang and Zhang, Jingbo and Lin, Haitao and Zhang, Wenyao and Wang, Jinghe and Jin, Hanglei and Xie, Shuzhao and Jiang, Jingyan and Wang, Zhi},
  journal={arXiv preprint arXiv:2608.06332},
  year={2026}
}

@inproceedings{bruce2024genie,
  title={Genie: Generative interactive environments},
  author={Bruce, Jake and Dennis, Michael D and Edwards, Ashley and Parker-Holder, Jack and Shi, Yuge and Hughes, Edward and Lai, Matthew and Mavalankar, Aditi and Steigerwald, Richie and Apps, Chris and others},
  booktitle={Forty-first international conference on machine learning},
  year={2024}
}

@inproceedings{ko2024learning,
  title={Learning to act from actionless videos through dense correspondences},
  author={Ko, Po-Chen and Mao, Jiayuan and Du, Yilun and Sun, Shao-Hua and Tenenbaum, Joshua B},
  booktitle={International Conference on Learning Representations},
  volume={2024},
  pages={40938--40958},
  year={2024}
}

@inproceedings{liu2025rdt,
  title={Rdt-1b: a diffusion foundation model for bimanual manipulation},
  author={Liu, Songming and Wu, Lingxuan and Li, Bangguo and Tan, Hengkai and Chen, Huayu and Wang, Zhengyi and Xu, Ke and Su, Hang and Zhu, Jun},
  booktitle={International Conference on Learning Representations},
  volume={2025},
  pages={29982--30009},
  year={2025}
}

@article{team2024octo,
  title={Octo: An open-source generalist robot policy},
  author={Team, Octo Model and Ghosh, Dibya and Walke, Homer and Pertsch, Karl and Black, Kevin and Mees, Oier and Dasari, Sudeep and Hejna, Joey and Kreiman, Tobias and Xu, Charles and others},
  journal={arXiv preprint arXiv:2405.12213},
  year={2024}
}

@article{cheang2024gr,
  title={Gr-2: A generative video-language-action model with web-scale knowledge for robot manipulation},
  author={Cheang, Chi-Lam and Chen, Guangzeng and Jing, Ya and Kong, Tao and Li, Hang and Li, Yifeng and Liu, Yuxiao and Wu, Hongtao and Xu, Jiafeng and Yang, Yichu and others},
  journal={arXiv preprint arXiv:2410.06158},
  year={2024}
}

@article{hu2024video,
  title={Video prediction policy: A generalist robot policy with predictive visual representations},
  author={Hu, Yucheng and Guo, Yanjiang and Wang, Pengchao and Chen, Xiaoyu and Wang, Yen-Jen and Zhang, Jianke and Sreenath, Koushil and Lu, Chaochao and Chen, Jianyu},
  journal={arXiv preprint arXiv:2412.14803},
  year={2024}
}

@article{su2024roformer,
  title={Roformer: Enhanced transformer with rotary position embedding},
  author={Su, Jianlin and Ahmed, Murtadha and Lu, Yu and Pan, Shengfeng and Bo, Wen and Liu, Yunfeng},
  journal={Neurocomputing},
  volume={568},
  pages={127063},
  year={2024},
  publisher={Elsevier}
}

@inproceedings{guo2026ctrl,
  title={Ctrl-world: A controllable generative world model for robot manipulation},
  author={Guo, Yanjiang and Shi, Lucy and Chen, Jianyu and Finn, Chelsea},
  booktitle={International Conference on Learning Representations},
  volume={2026},
  pages={6121--6138},
  year={2026}
}

@article{gao2026dreamdojo,
  title={Dreamdojo: A generalist robot world model from large-scale human videos},
  author={Gao, Shenyuan and Liang, William and Zheng, Kaiyuan and Malik, Ayaan and Ye, Seonghyeon and Yu, Sihyun and Tseng, Wei-Cheng and Dong, Yuzhu and Mo, Kaichun and Lin, Chen-Hsuan and others},
  journal={arXiv preprint arXiv:2602.06949},
  year={2026}
}

@inproceedings{liao2026genie,
  title={Genie envisioner: A unified world foundation platform for robotic manipulation},
  author={Liao, Yue and Zhou, Pengfei and Huang, Siyuan and Yang, Donglin and Chen, Shengcong and Jiang, Yuxin and Hu, Yue and Liu, Si and Luo, Jianlan and Chen, Liliang and others},
  booktitle={International Conference on Learning Representations},
  volume={2026},
  pages={88446--88463},
  year={2026}
}

@article{khazatsky2024droid,
  title={Droid: A large-scale in-the-wild robot manipulation dataset},
  author={Khazatsky, Alexander and Pertsch, Karl and Nair, Suraj and Balakrishna, Ashwin and Dasari, Sudeep and Karamcheti, Siddharth and Nasiriany, Soroush and Srirama, Mohan Kumar and Chen, Lawrence Yunliang and Ellis, Kirsty and others},
  journal={arXiv preprint arXiv:2403.12945},
  year={2024}
}

@article{huang2026pointworld,
  title={Pointworld: Scaling 3d world models for in-the-wild robotic manipulation},
  author={Huang, Wenlong and Chao, Yu-Wei and Mousavian, Arsalan and Liu, Ming-Yu and Fox, Dieter and Mo, Kaichun and Fei-Fei, Li},
  journal={arXiv preprint arXiv:2601.03782},
  year={2026}
}

@article{du2023learning,
  title={Learning universal policies via text-guided video generation},
  author={Du, Yilun and Yang, Sherry and Dai, Bo and Dai, Hanjun and Nachum, Ofir and Tenenbaum, Josh and Schuurmans, Dale and Abbeel, Pieter},
  journal={Advances in neural information processing systems},
  volume={36},
  pages={9156--9172},
  year={2023}
}

@misc{li2025unifiedvideoactionmodel,
      title={Unified Video Action Model}, 
      author={Shuang Li and Yihuai Gao and Dorsa Sadigh and Shuran Song},
      year={2025},
      eprint={2503.00200},
      archivePrefix={arXiv},
      primaryClass={cs.RO},
      url={https://arxiv.org/abs/2503.00200}, 
}

@article{zhu2025unified,
  title={Unified world models: Coupling video and action diffusion for pretraining on large robotic datasets},
  author={Zhu, Chuning and Yu, Raymond and Feng, Siyuan and Burchfiel, Benjamin and Shah, Paarth and Gupta, Abhishek},
  journal={arXiv preprint arXiv:2504.02792},
  year={2025}
}

@article{kim2026cosmos,
  title={Cosmos policy: Fine-tuning video models for visuomotor control and planning},
  author={Kim, Moo Jin and Gao, Yihuai and Lin, Tsung-Yi and Lin, Yen-Chen and Ge, Yunhao and Lam, Grace and Liang, Percy and Song, Shuran and Liu, Ming-Yu and Finn, Chelsea and others},
  journal={arXiv preprint arXiv:2601.16163},
  year={2026}
}

@article{ye2026world,
  title={World action models are zero-shot policies},
  author={Ye, Seonghyeon and Ge, Yunhao and Zheng, Kaiyuan and Gao, Shenyuan and Yu, Sihyun and Kurian, George and Indupuru, Suneel and Tan, You Liang and Zhu, Chuning and Xiang, Jiannan and others},
  journal={arXiv preprint arXiv:2602.15922},
  year={2026}
}

@article{li2026wall,
  title={WALL-WM: Carving World Action Modeling at the Event Joints},
  author={Li, Shalfun and Yao, Victor and Yang, Charles and Qu, Truth and Cheng, Regis and Yu, Ryan and Lu, Howard and Von, Newton and Chen, Vincent and Tang, Yohann and others},
  journal={arXiv preprint arXiv:2606.01955},
  year={2026}
}

@article{yang20264d,
  title={4D-WAM: Infusing Spatiotemporal Awareness into World Action Models through Trajectory Fields},
  author={Yang, Lishan and Song, Wenxuan and Wang, Xi and Sheng, Pingyue and Fang, Zheng and Zhou, Ziyang and He, Junjie and Yan, Haodong and Chen, Jiayi and Sun, Nan and others},
  journal={arXiv preprint arXiv:2608.08023},
  year={2026}
}

@article{chen2026bridgev2w,
  title={Bridgev2w: Bridging video generation models to embodied world models via embodiment masks},
  author={Chen, Yixiang and Li, Peiyan and Yang, Jiabing and He, Keji and Wu, Xiangnan and Xu, Yuan and Wang, Kai and Liu, Jing and Liu, Nianfeng and Huang, Yan and others},
  journal={arXiv preprint arXiv:2602.03793},
  year={2026}
}

@article{zhen2026action,
  title={Action Images: End-to-End Policy Learning via Multiview Video Generation},
  author={Zhen, Haoyu and Gao, Zixian and Sun, Qiao and Zhao, Yilin and Yang, Yuncong and Du, Yilun and Guo, Pengsheng and Wang, Tsun-Hsuan and Qiao, Yi-Ling and Gan, Chuang},
  journal={arXiv preprint arXiv:2604.06168},
  year={2026}
}

@misc{li2026spatialvamspatialawaremultiviewvideodiffusion,
      title={SpatialVAM:Spatial-Aware Multi-View Video Diffusion as a Data-Efficient Robot Policy}, 
      author={Peiyan Li and Yixiang Chen and Yuan Xu and Jiabing Yang and Xiangnan Wu and Jun Guo and Nan Sun and Long Qian and Xinghang Li and Xin Xiao and Jing Liu and Nianfeng Liu and Tao Kong and Yan Huang and Liang Wang and Tieniu Tan},
      year={2026},
      eprint={2604.03181},
      archivePrefix={arXiv},
      primaryClass={cs.RO},
      url={https://arxiv.org/abs/2604.03181}, 
}

@article{liu2022flow,
  title={Flow straight and fast: Learning to generate and transfer data with rectified flow},
  author={Liu, Xingchao and Gong, Chengyue and Liu, Qiang},
  journal={arXiv preprint arXiv:2209.03003},
  year={2022}
}

@article{liu2026triworldbench,
  title   = {TriWorldBench: A Tri-View Consistency Perspective on Embodied World Models},
  author  = {Liu, Xuanyi and Wang, Haofeng and Li, Ruiqi and Yu, Danni and Wan, Rui and Zhang, Ruixu and Tao, Siyu and Yang, Xue and Zhang, Shaofeng and Zhang, Zicheng and Zhang, Jiaqi and Ma, Siwei},
  journal = {arXiv preprint arXiv:2609.26314},
  year    = {2026}
}

@article{wang2026world,
  title={World action models: The next frontier in embodied ai},
  author={Wang, Siyin and Shi, Junhao and Fu, Zhaoyang and He, Xinzhe and Liu, Feihong and Yang, Chenchen and Zhou, Yikang and Fei, Zhaoye and Gong, Jingjing and Fu, Jinlan and others},
  journal={arXiv preprint arXiv:2605.12090},
  year={2026}
}

@article{wen2026dependency,
  title={Dependency, Compression, and Synergy: A Unified Information-Theoretic View of Multimodal Learning},
  author={Wen, Liangjian and Li, Linjie and Duan, Jiang and Dai, Yong and Liu, Jianzhuang and Kang, Zhao},
  journal={arXiv preprint arXiv:2609.14421},
  year={2026}
}

@article{zhang2026pelican,
  title={Pelican-Unify 1.0: A Unified Embodied Intelligence Model for Understanding, Reasoning, Imagination and Action},
  author={Zhang, Yi and Chen, Yinda and Liu, Che and Ding, Zeyuan and Xu, Jin and Zou, Shilong and Liao, Junwei and Hu, Jiayu and Ren, Xiancong and Zhang, Xiaopeng and others},
  journal={arXiv preprint arXiv:2605.15153},
  year={2026}
}

@inproceedings{shi-etal-2026-vla,
    title = "{VLA}-Trace: Diagnosing Vision-Language-Action Models through Representation and Behavior Tracing",
    author = "Shi, Haoyuan  and
      Ren, Xiancong  and
      Zhang, Yingji  and
      Zhang, Qinfan  and
      Hu, Jiayu  and
      Shan, Haozhe  and
      Dong, Han  and
      Lu, Jinpeng  and
      Chen, Yinda  and
      Zhang, Yi  and
      Dai, Yong  and
      Ju, Xiaozhu",
    editor = "Lee, Hung-Yi  and
      Dev, Sunipa  and
      Li, Junyi Jessy  and
      Elliott, Desmond",
    booktitle = "Proceedings of the 2026 Conference on Empirical Methods in Natural Language Processing",
    month = oct,
    year = "2026",
    address = "Budapest, Hungary",
    publisher = "Association for Computational Linguistics",
    url = "https://aclanthology.org/2026.emnlp-main.2076/",
    doi = "10.18653/2026.emnlp-main.2076",
    pages = "47280--47315",
    ISBN = "979-8-89176-459-0"
}

@misc{luo2026roboexpandreal2sim2realdataengine,
      title={RoboExpand: A Real2Sim2Real Data Engine with Agentic Reconstruction and Multi-View Video Generation},
      author={Yifan Luo and Xin Wu and Xiancong Ren and Chenhui Deng and Huimin Ma and Jianxin Sun and Yifan Sun and Qinfan Zhang and Han Dong and Jin Xu and Zhiheng Liu and Chen Sun and Yong Dai and Bing Su and Xiaozhu Ju},
      year={2026},
      eprint={2611.36418},
      archivePrefix={arXiv},
      primaryClass={cs.RO},
      url={https://arxiv.org/abs/2611.36418},
}

@article{wang2026rethinking,
  title={Rethinking visual-language-action model scaling: Alignment, mixture, and regularization},
  author={Wang, Ye and Zheng, Sipeng and Luo, Hao and Zhang, Wanpeng and Yuan, Haoqi and Xu, Chaoyi and Xu, Haiweng and Feng, Yicheng and Yu, Mingyang and Kang, Zhiyu and others},
  journal={arXiv preprint arXiv:2602.09722},
  year={2026}
}

@article{bai2025qwen3,
  title={Qwen3-vl technical report},
  author={Bai, Shuai and Cai, Yuxuan and Chen, Ruizhe and Chen, Keqin and Chen, Xionghui and Cheng, Zesen and Deng, Lianghao and Ding, Wei and Gao, Chang and Ge, Chunjiang and others},
  journal={arXiv preprint arXiv:2511.21631},
  year={2025}
}
\bibliographystyle{plain}

\appendix
\clearpage
\section{Appendix}

\begin{figure}[h]
    \centering
    \includegraphics[width=0.45\linewidth]{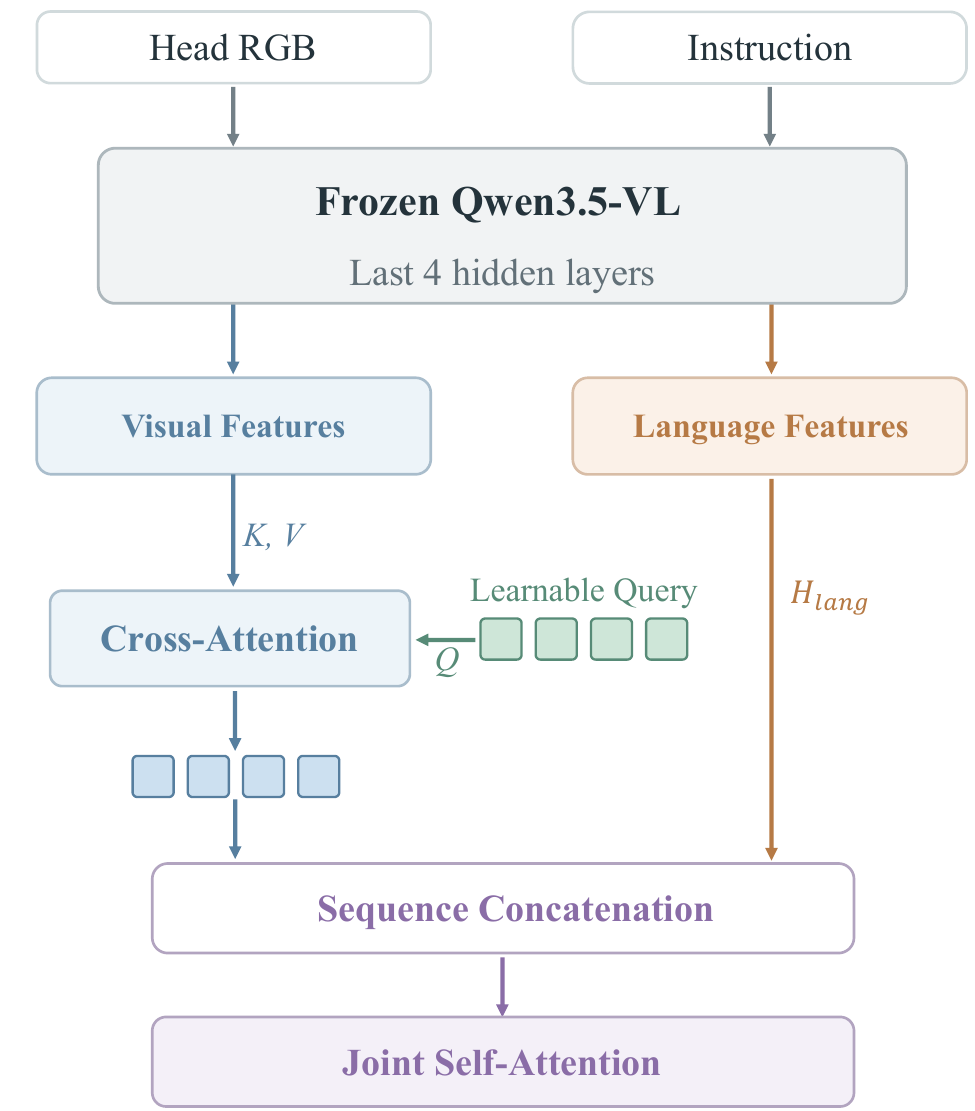}
    \caption{Trainable adapter modulates main view and instruction information.}
    \label{fig:adapter}
\end{figure}

\begin{table}[h]
    \centering
    \small
    \caption{Training and inference hyperparameters.}
    \begin{tabular}{lc}
        \toprule
        \textbf{Hyperparameter} & \textbf{Value} \\
        \midrule
        Training Resource & $16\times\text{Nvidia B200}$ \\
        Batch Size & 16 \\
        Optimizer & AdamW \\
        Weight Decay & $1\times10^{-2}$ \\
        Learning Rate & $1\times10^{-5}$ \\
        Learning Rate Scheduler & Cosine Scheduler \\
        Learning Rate Warmup Steps & 1000 \\
        Simulation Action Chunk & 40 \\
        Real-World Action Chunk & 80 \\
        Main view Resolution & $640\times480$ \\
        RGB Sequence Resolution & $320\times224$ \\
        Action-View Sequence Resolution & $320\times224$ \\
        Forward Dynamics Sample Probability & 0.2 \\
        Inverse Dynamics Sample Probability & 0.2 \\
        joint Generation Sample Probability & 0.6 \\
        Image-to-Video Sample Probability & 0.7 \\
        Video-to-Video Sample Probability & 0.3 \\
        Inference Denoising Steps & 8 \\
        \bottomrule
    \end{tabular}
    \label{tab:training_params}
\end{table}

\subsection{Implementation Details.}
\label{app:implementation_details}

\paragraph{Video encoding and decoding.}
We use Wan2.2VAE~\cite{wan2025wan} encoder $\mathcal E$ and decoder $\mathcal D$, shared across RGB and action-view sequences. Each sequence is encoded separately, with a spatial downsampling factor of 16 and a temporal downsampling factor of 4. A $T$-frame sequence has the latent shape
\begin{equation}
    y^s\in\mathbb R^{3\times T\times H\times W}
    \xrightarrow{\;\mathcal E\;}
    z_s\in\mathbb R^{48\times L\times(H/16)\times(W/16)},
    \quad
    L=1+\frac{T-1}{4},
    \label{eq:app:vae-shape}
\end{equation}
for $T=4n+1$ and spatial dimensions divisible by 16. The initial frame is encoded separately. At the sequence resolution of $320\times224$ pixels (width $\times$ height), each latent grid has spatial size $14\times20$. For the single-frame-conditioned window in Method, we set $T=k+1$: the simulation horizon $k=40$ and real-world horizon $k=80$. Generated latents are decoded sequence-wise as $\hat y^s=\mathcal D(\hat z_s)$ before RGB evaluation or action recovery. The $640\times480$ main-view input to the semantic branch is separate from these lower-resolution video sequences.

\paragraph{Training-mode sampling.}
Table~\ref{tab:training_params} summarizes the optimization settings, resolutions, horizons, and sampling probabilities. For each training example, we first select forward dynamics, inverse dynamics, or joint generation with probabilities 0.2, 0.2, and 0.6. These are operating modes of one shared model. Following Table~\ref{tab:schedule}, target sequence set is $M=\mathcal S_{\mathrm{rgb}}$, $M=V$, or $M=\mathcal S$, respectively, sequences outside $M$ remain clean for conditions. We then sample image-to-video (I2V) or video-to-video (V2V) conditioning with probabilities 0.7 and 0.3.

\begin{table*}[t]
    \caption{Mode-specific generation choices. Frame $f=0$ always remains clean.}
    \label{tab:schedule}
    \centering
    \begin{tabular}{lccc}
    \toprule
    \textbf{Mode} & $M$ & \textbf{RGB sequences ($f \geq 1$)} & \textbf{Action views ($f \geq 1$)} \\
    \midrule
    Forward dynamics  & $\mathcal{S}_{\mathrm{rgb}}$ & noised & clean  \\
    Inverse dynamics  & $V$                          & clean  & noised \\
    Joint generation  & $\mathcal{S}$                & noised & noised \\
    \bottomrule
    \end{tabular}
\end{table*}

\paragraph{Temporal-prefix conditioning.}
Here, I2V and V2V specify the clean temporal prefix of the sequences selected for generation. I2V preserves only the initial frame, whereas V2V preserves a longer prefix and predicts the remaining segment. We set one fifth of a simulation training clip and one quarter of a real-world training clip for V2V prefix. Let $L_c$ be the number of clean prefix latents, the training mask generalizes to
\begin{equation}
    m_{s,f}=\mathbbm{1}[s\in M\wedge f\geq L_c],
    \qquad 0\leq f<L,
    \label{eq:app:prefix-mask}
\end{equation}
with $L_c=1$ for I2V and $L_c>1$ for V2V. The loss remains restricted to corrupted positions. All closed-loop results reported in the experiments use I2V mode.

\subsection{Rendering Details.}
\label{app:rendering}

\paragraph{URDF-based rendering.}
Forward kinematics maps $S_t$ and $A_{t+1:t+k}$ to current and future action views. We render the URDF-connected \texttt{base\_link}, arm-joint nodes, and two gripper nodes at $(W,H)=(320,224)$ pixels. Base-to-arm links are yellow, arm nodes blue, and gripper colors vary from green (open) to red (closed). Camera-space depth modulates color intensity while preserving these roles.

\paragraph{Task-level camera rig.}
For each dataset-embodiment-task training group, we save one \texttt{task\_rig.json}, shared by its episodes. Four virtual cameras have the ordered roles \emph{front}, \emph{top}, \emph{left}, and \emph{right}, no physical camera is used. Extrinsics are fixed first: each camera looks at the mean arm-root position over the group's initial frames, at distance \texttt{dist}.

\paragraph{Intrinsic selection.}
For view $i$, let $\mathcal P_i$ contain camera-space joint points from all valid frames of all training episodes in the group, discarding points with $Z\leq10^{-4}$. With occupancy $o$ (default $0.95$), the union determines
\begin{equation}
h_{w,i}=\max_{(X,Y,Z)\in\mathcal P_i}|X/Z|,
\quad h_{h,i}=\max_{(X,Y,Z)\in\mathcal P_i}|Y/Z|,\quad
f_i=\min\!\left(\frac{oW}{2h_{w,i}},\frac{oH}{2h_{h,i}}\right).
\end{equation}
Each view has its own $f_i$, with square pixels ($f_{x,i}=f_{y,i}=f_i$), zero skew, and fixed principal point $(W/2,H/2)$. Thus $(u,v)=(f_iX/Z+W/2,f_iY/Z+H/2)$. Maxima are taken over points, frames, and episodes, without averaging. The occupancy bound applies along both image dimensions about the optical axis. For asymmetric trajectories, the farther side determines the focal length, leaving more margin on the opposite side.

If $\mathcal P_i$ is empty or either extent is near zero, we use $f_i=(H/2)/\tan(\phi/2)$ with vertical field of view $\phi=60^\circ$. Changing \texttt{dist} changes camera positions; focal lengths are recomputed to maintain the occupancy bound.

\paragraph{Consistent rendering at inference.}
Saved intrinsics and extrinsics remain fixed across frames and episodes and are reused for conditioning views and action recovery. Test trajectories do not determine the preset. View roles, ordering, resolution, and appearance rules are shared across embodiments, numerical camera parameters are group-specific.

\subsection{Action Recovery Mechanism.}
\label{app:recovery}

\paragraph{Overview.}
Recovery combines frame-wise estimation, window-level refinement, and gripper decoding. At local frame $j=0,\ldots,k$, corresponding to time $t+j$, $u_j=(q_j,g_j)$ contains arm configurations and gripper variables. We anchor $u_0=S_t$ without using future ground-truth states. The URDF, virtual-camera preset, and rendering conventions match the training action views.

\paragraph{Stage A: frame-wise estimation.}
The experimental GPU path refines 32 initializations per frame with batched Levenberg--Marquardt (LM). Neighboring-frame solutions supply additional initializations in Jacobi-style rounds. Anchor frames are solved at stride $s_{\mathrm{anchor}}=3$; intermediate configurations are interpolated and batch-refined, and high-cost segments are revisited. An independent forward chain propagates estimates from the observed frame. Disagreement among restart solutions with similar costs provides the \emph{spread} signal for Stage B.

\paragraph{Gauss-Newton refinement.}
The \texttt{ridge} option extracts medial-axis information from predicted RGB action views. Multiview image-field residuals define the local least-squares problem. With residual vector $r_j$, Jacobian $J_j=\partial r_j/\partial u_j$, and residual weights $W_j$, the damped normal equations are
\begin{equation}
    \bigl(J_j^\top W_jJ_j+\mu M\bigr)\Delta u_j
    =-J_j^\top W_jr_j,
    \label{eq:app:recovery-gn}
\end{equation}
where $M$ is a positive-definite damping metric. Setting $\mu=0$ gives Gauss-Newton, positive damping stabilizes the LM step. Coarse-to-fine field scales guide image alignment.

\paragraph{Stage B: window-level estimation.}
Given frame-wise estimates $\hat u_j$, we fit a cubic B-spline trajectory $z_j=(BC)_{j,:}^{\top}$ with knots every four frames:
\begin{equation}
    \min_C\;
    \sum_{j=0}^{k}(z_j-\hat u_j)^\top
        \widetilde H_j(z_j-\hat u_j)
    +\lVert D^{(2)}BC\Lambda^{1/2}\rVert_F^2,
    \label{eq:app:recovery-map}
\end{equation}
Here $B$ is the temporal basis, $C$ contains spline coefficients, and $D^{(2)}$ is the second-difference operator. Diagonal data weights $\widetilde H_j$ use the diagonal of $J_j^\top W_jJ_j$, attenuated for large Stage A spread. Diagonal $\Lambda$ controls coordinate-wise smoothness, with stronger regularization for each arm's sixth joint in the supplied bimanual configuration (\texttt{lam\_joint6}=5). The fit balances image evidence against temporal regularization, which is part of estimation.

With \texttt{n\_outer}=2, the first window fit uses Stage A estimates. A single-start refinement at finer image-field scales updates the estimates and curvature weights for the second fit, retaining Stage A spread. After each fit, configurations are clipped to joint limits and frame 0 is restored to $S_t$.

\paragraph{Gripper decoding.}
After Stage B, the renderer's red-green fingertip encoding replaces gripper estimates with color-decoded values. A three-frame median filter is applied, and only missing runs of at most three frames are interpolated.

\paragraph{Results Evaluation.}
After gripper decoding, for each recovered configuration $A\in\mathcal Q(U)$, where $\mathcal Q(U)$ encodes the URDF joint limits, it is rendered as $x^i(A)=R(\operatorname{FK}(A;U);K_i,T_i)$ and scored by
\begin{equation}
E_{j}(A)
  = \sum_{i \in V}
        \Big[\, 1 - \mathrm{IoU}\big(x^{i}(A),\, \tilde{x}^{i}_{j}\big) \Big]
     + \lambda \sum_{i \in V}
        d^{\rightarrow}\big(\tilde{x}^{i}_{j},\, x^{i}(A)\big),~\lambda\geq{0}.
\label{eq:recovery-cost}
\end{equation}
Here, IoU measures silhouette overlap, the one-sided Chamfer term $d^{\rightarrow}(\cdot)$ measures predicted-to-rendered foreground distance, including disjoint silhouettes.

\subsection{Real-World Experiment Details.}
\label{app:real-world}

\begin{figure}[t]
    \centering
    \vspace{0.5em}
    \includegraphics[width=1.0\linewidth]{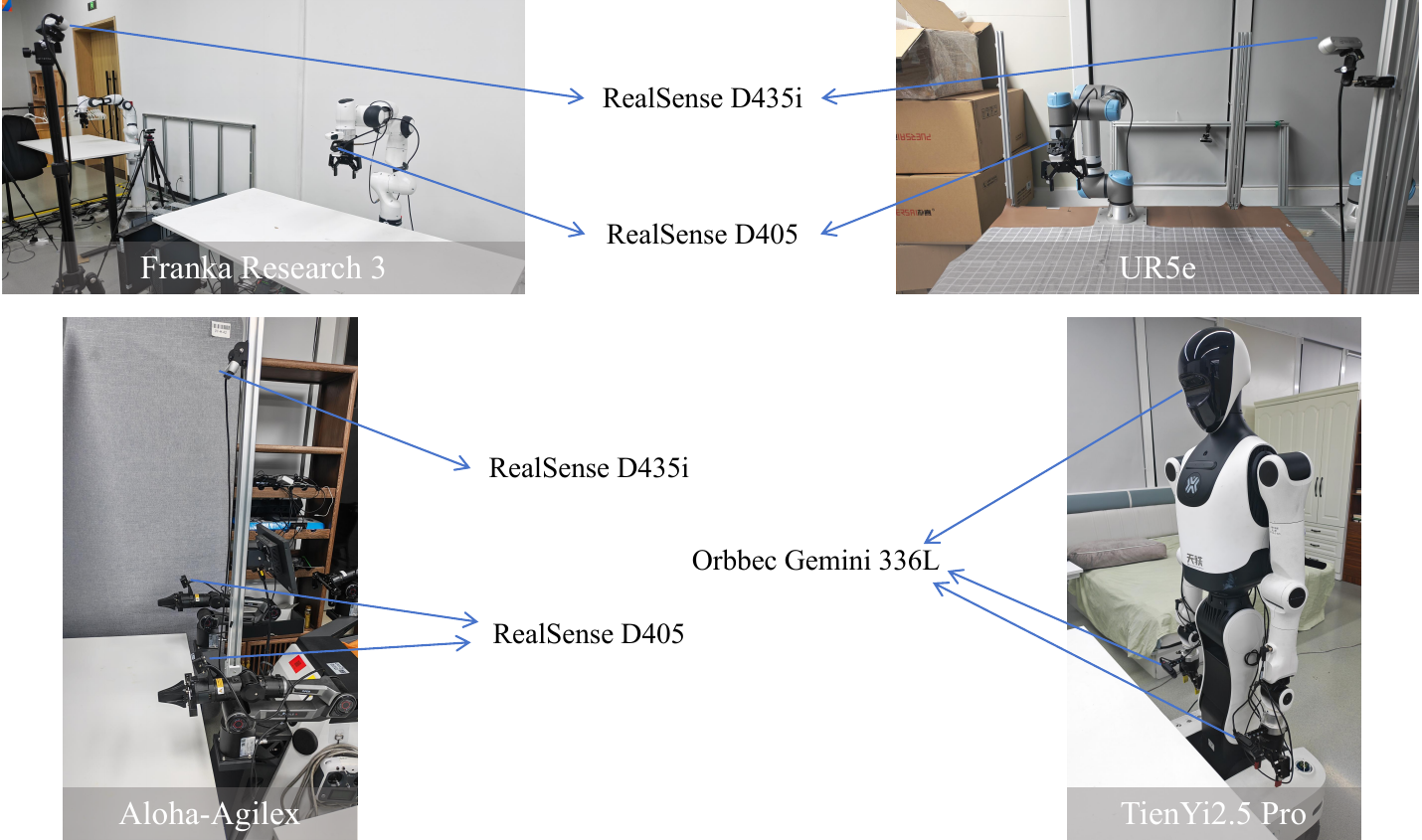}
    \caption{Real-world robots configuration. Franka and UR use one front camera and one wrist camera, while Aloha-Agilex and TienYi use one head camera and two wrist cameras.}
    \vspace{-0.5em}
    \label{fig:setup}
\end{figure}

\paragraph{Task design.}
All four tasks span target selection, tool-mediated contact, bimanual coordination, and sequential object interaction. As shown in Section~\ref{sec:experiments:real_world}, all four platforms are evaluated on the first two tasks, and only the bimanual platforms are evaluated on the remaining two tasks:

\begin{enumerate}[leftmargin=*,nosep]
    \item \textbf{Place\_block}. A box and several small blocks are placed on the table, with randomized block positions. The robot must select and grasp the specific blue block, place it inside the box, and then return to its initial pose.
    \item \textbf{Wipe\_plate}. A plate containing dirty water is placed on the table, with a sponge initially positioned to its right. The robot must grasp the sponge, wipe the plate clean, and place the sponge on the opposite side of the plate.
    \item \textbf{Stack\_blocks}. A yellow block is placed on the left side of the table, and the red and blue blocks are placed on the right, with positions randomized within these regions. The robot must coordinate both arms to stack three blocks, ordered red, yellow, and blue from bottom to top.
    \item \textbf{Storage\_item}. A drawer and a block are placed on the table. The left arm first opens the drawer, the right arm then grasps the block and places it inside, finally, the left arm closes the drawer.
\end{enumerate}

\paragraph{Settings.}
Figure~\ref{fig:setup} illustrates the physical setups. Franka Research 3 and UR5e each use one external RealSense D435i camera, one wrist-mounted RealSense D405 camera, and a Robotiq gripper. Aloha-Agilex uses a head-mounted RealSense D435i camera, two wrist-mounted RealSense D405 cameras, and two Robotiq grippers. TienYi2.5 Pro uses Orbbec Gemini 336L cameras for all three viewpoints (one head and two wrists), together with two Robotiq grippers. The two or three RGB sequences are encoded separately and processed by sequence-specific RoPE offsets. This input formulation supports joint training across embodiments with different numbers of observation views using a single shared model.

\begin{figure*}
    \centering
    \includegraphics[width=\linewidth]{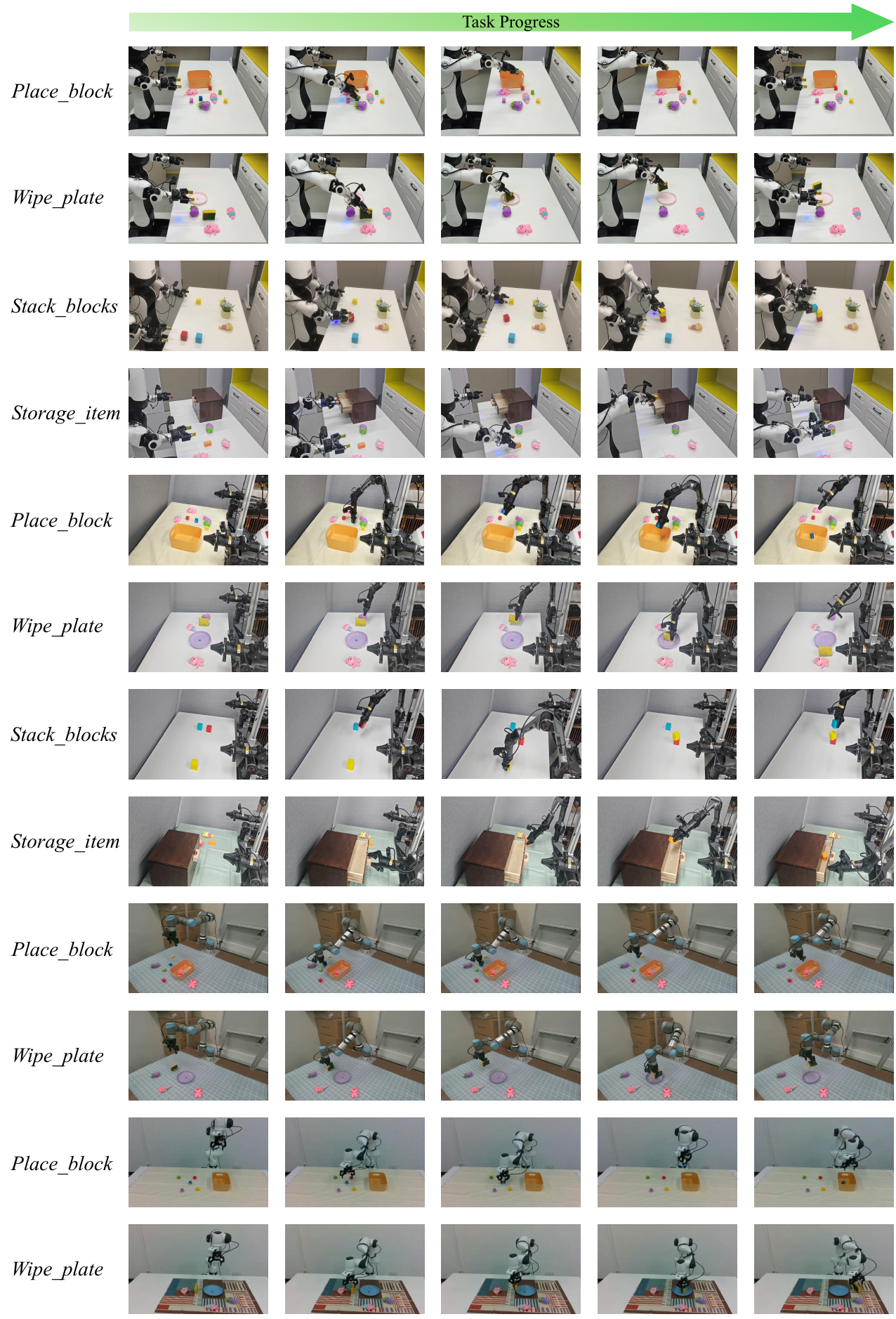}
    \caption{Execution progress of real-world manipulation tasks. Each row shows representative temporal snapshots for one task, illustrating how the robot completes place\_block, wipe\_plate, stack\_blocks, and storage\_item.}
    \label{fig:tasks}
\end{figure*}

\paragraph{Inference configuration.}
All four platforms use the same jointly trained checkpoint, with their corresponding URDFs and saved virtual-camera presets used for rendering and recovery. During inference, our model predicts an 80-step action chunk ($k=80$), encoding target joint configurations and gripper states as action views for recovery into executable commands. In our experimental setup with an NVIDIA A800 GPU, inference and action recovery take approximately 6.9 s and 2.7 s per chunk, respectively. We execute each predicted action chunk in full. After execution, we obtain new observations and use them to plan the next action chunk. Figure~\ref{fig:tasks} shows each embodiment-task execution progress.

\begin{table*}[t]
    \centering
    \small
    \caption{Multi-view action recovery errors across five embodiments on 31 shared RoboTwin 2.0 tasks. Recovery errors use GT action views with five held-out trajectories per task and 40 future frames per window, aggregated with equal task weighting.}
    \label{tab:embodiment_recovery_analysis}
    \begin{tabular}{lcccccccccc}
        \toprule
        & \multicolumn{5}{c}{Position (mm) $\downarrow$}
        & \multicolumn{5}{c}{Rotation ($^\circ$) $\downarrow$} \\
        \cmidrule(lr){2-6}
        \cmidrule(lr){7-11}
        Embodiment
        &  One
        &  Two
        &  Three
        & Four
        &  Five
        &  One
        &  Two
        &  Three
        & Four
        &  Five \\
        \midrule
        Aloha-Agilex
        &  82.85
        &  4.63
        &  1.91
        & 1.39
        &  \textbf{1.23}
        &  38.36
        &  11.59
        &  3.07
        & \textbf{1.57}
        &  1.60 \\
        Piper
        &  147.61
        &  11.99
        &  3.28
        & \textbf{1.66}
        &  1.93
        &  60.37
        &  17.54
        &  3.72
        & \textbf{1.71}
        &  1.74 \\
        ARX-X5
        &  143.37
        &  6.00
        &  2.83
        & \textbf{1.22}
        &  1.35
        &  54.52
        &  16.19
        &  3.33
        & \textbf{1.95}
        &  2.37 \\
        Franka-Panda
        &  165.89
        &  12.14
        &  7.30
        & 4.11
        &  \textbf{3.98}
        &  56.30
        &  22.54
        &  13.25
        & 9.45
        &  \textbf{9.13} \\
        UR5-Xsg
        &  60.51
        &  22.71
        &  17.75
        & \textbf{11.85}
        &  11.88
        &  46.80
        &  33.31
        &  26.85
        & 21.39
        &  \textbf{21.17} \\
        \bottomrule
    \end{tabular}
\end{table*}
\begin{table*}[h]
    \centering
    \small
    \caption{Comparison between an Aloha-only checkpoint and the multi-embodiment checkpoint trained on the same 31 tasks. Displayed tasks are selected based on success-rate gaps exceeding 15 percentage points between the originally reported results in Tables~\ref{tab:robotwin_full} and~\ref{tab:robotwin_five_embodiments}. Here we report success rates over 25 trials per task under randomized settings.}
    \label{tab:aloha_simulated_comparison}
    \begin{tabular}{lccc}
        \toprule
        Task & Single Checkpoint & Joint Checkpoint & $\Delta$ (pp) \\
        \midrule
        \textit{Blocks Ranking RGB} & 20\% &  0\% & $+20$ \\
        \textit{Move Can Pot}       & 56\% & 72\% & $-16$ \\
        \textit{Move Stapler Pad}   & 72\% & 96\% & $-24$ \\
        \textit{Open Laptop}        & 44\% & 96\% & $-52$ \\
        \textit{Place Bread Basket} & 56\% & 68\% & $-12$ \\
        \textit{Stack Blocks Two}   & 28\% &  4\% & $+24$ \\
        \midrule
    \end{tabular}
    \vspace{-1.0em}
\end{table*}

\subsection{Multi-Embodiment Experiment Analysis.}
\label{app:multi_embodiments}

\paragraph{Evaluation protocol.}\quad
We evaluate action recovery using five held-out trajectories per task on the 31 shared RoboTwin~2.0 tasks. The solver recovers joint configurations from ground-truth (GT) action views, and forward kinematics gives the corresponding end-effector poses. Let $(\hat{\mathbf p},\hat{\mathbf R})$ and $(\mathbf p,\mathbf R)$ denote the recovered and GT poses, with positions expressed in millimeters. We compute
\begin{equation}
\begin{aligned}
e_{\mathrm{pos}}
&= \|\hat{\mathbf p}-\mathbf p\|_2,
\\
e_{\mathrm{rot}}
&= \frac{180}{\pi}\arccos\!\left(
\operatorname{clip}\!\left(
\frac{\operatorname{tr}(\mathbf R^\top\hat{\mathbf R})-1}{2},
-1,1\right)\right).
\end{aligned}
\end{equation}
Position and rotation errors are reported in millimeters and degrees, respectively. Rotation error measures the relative end-effector rotation angle. Each recovery window contains one known initial frame and 40 future frames. We take the larger error across arms at each frame, average over each complete trajectory, and then equally average over trajectories and tasks. Overlapping frames are counted only once, and the initial frame is excluded from scoring.

\paragraph{Results and analysis.}
Table~\ref{tab:embodiment_recovery_analysis} reports GT-view recovery errors. Combined with the closed-loop success rate in Table~\ref{tab:robotwin_multi_embodiment}, the lower success rates of Franka-Panda and UR5-Xsg should be interpreted in the context of their simulated bimanual configurations, assembled from two single-arm robots rather than designed as integrated dual-arm systems. Limitations in this integration may affect inter-arm coordination and joint-target tracking, introducing execution effects not captured by the URDF-based kinematic model used for recovery. Consequently, even accurately recovered joint targets may not be realized precisely by the simulated system, affecting grasping and placement. The lower demonstration-collection success rates observed for these configurations are consistent with such execution challenges. Together with the measured recovery errors, these configuration-level factors may contribute to their lower closed-loop success rates.

We additionally evaluate action-recovery accuracy by progressively adding views in the order Front, Left, Top, Right and Back. The results show that more views do not necessarily improve recovery. Our default four-view configuration already provides strong geometric constraints, whereas adding poorly informative views can interfere with the constraints from more informative views during joint optimization, reducing recovery accuracy.

\subsection{Joint-Checkpoint Results Analysis.}
To investigate the substantial success-rate drops on certain tasks under joint training, we train an Aloha-Agilex-only model for 20k steps with a batch size of 16 on the same 31 tasks, achieving an average success rate of 69.68\%. Then we compare the single and joint checkpoint on the six tasks within this set whose success rate gaps exceed 15 percentage points in the originally reported results.

As shown in Table~\ref{tab:aloha_simulated_comparison}, with the task set held fixed, this comparison suggests that multi-embodiment joint training may introduce task-specific interference. This interpretation is consistent with negative transfer reported for heterogeneous robot-data mixtures by~\citep{wang2026rethinking}, where adding diverse robot data does not necessarily improve performance uniformly across tasks.

\subsection{Limitations and Future Work.}
\label{app:limitations}
The demonstrated scope of Dream4ACT is bounded by its training data and embodiment coverage. Our evaluation includes five simulated embodiments and four real-world platforms, broader task coverage and transfer to unseen embodiments remain to be validated. Kinematic mismatch can affect physical execution, while joint-recovery accuracy also depends on generated-view fidelity. Explicit multiview generation and action recovery incur substantial latency: in our real-world setup, inference and recovery take approximately 6.9\,s and 2.7\,s, respectively, per 80-step chunk. The current implementation therefore supports chunk-based execution rather than real-time, per-step replanning. Expanding demonstration and embodiment coverage, improving robustness to kinematic mismatch, and accelerating generation and recovery are promising directions for extending this shared visual action interface in the future. We also plan to incorporate large-scale data through pretraining or mid-training to improve generalization to unseen tasks and embodiments.

\begin{table*}[t]
\centering
    \caption{Per-task success rates (\%) on the 50-task RoboTwin 2.0 benchmark under clean and randomized (Rand.) settings. Baseline scores are taken from publicly reported results. Bold marks the highest score for each task and setting.}
\label{tab:robotwin_full}
\small
\setlength{\tabcolsep}{3pt}
\resizebox{\textwidth}{!}{%
\begin{tabular}{l cc cc cc cc cc}
\toprule
\multirow{2}{*}{\textbf{Simulation Task}} &
\multicolumn{2}{c}{\shortstack{$\pi_{0.5}$}} &
\multicolumn{2}{c}{\shortstack{Motus}} &
\multicolumn{2}{c}{\shortstack{LingBot-VA}} &
\multicolumn{2}{c}{\shortstack{Fast-WAM}} &
\multicolumn{2}{c}{\textbf{Ours}} \\
\cmidrule(lr){2-3} \cmidrule(lr){4-5} \cmidrule(lr){6-7} \cmidrule(lr){8-9} \cmidrule(lr){10-11}
 & Clean & Rand. & Clean & Rand. & Clean & Rand. & Clean & Rand. & Clean & Rand. \\
\midrule
\textit{Adjust Bottle} & \textbf{100\%} & 99\% & 89\% & 93\% & 90\% & 94\% & \textbf{100\%} & \textbf{100\%} & 91\% & 97\% \\
\textit{Beat Block Hammer} & 96\% & 93\% & 95\% & 88\% & 96\% & \textbf{98\%} & 99\% & 97\% & \textbf{100\%} & 96\% \\
\textit{Blocks Ranking RGB} & 92\% & 85\% & 99\% & 97\% & 99\% & 98\% & \textbf{100\%} & \textbf{100\%} & 96\% & 90\% \\
\textit{Blocks Ranking Size} & 49\% & 26\% & 75\% & 63\% & \textbf{94\%} & 96\% & \textbf{94\%} & \textbf{98\%} & 62\% & 58\% \\
\textit{Click Alarmclock} & 98\% & 89\% & \textbf{100\%} & \textbf{100\%} & 99\% & \textbf{100\%} & \textbf{100\%} & \textbf{100\%} & \textbf{100\%} & \textbf{100\%} \\
\textit{Click Bell} & 99\% & 66\% & \textbf{100\%} & \textbf{100\%} & \textbf{100\%} & \textbf{100\%} & \textbf{100\%} & \textbf{100\%} & \textbf{100\%} & \textbf{100\%} \\
\textit{Dump Bin Bigbin} & 92\% & \textbf{97\%} & 95\% & 91\% & 89\% & 96\% & \textbf{97\%} & 96\% & 90\% & 93\% \\
\textit{Grab Roller} & \textbf{100\%} & \textbf{100\%} & \textbf{100\%} & \textbf{100\%} & \textbf{100\%} & \textbf{100\%} & \textbf{100\%} & \textbf{100\%} & \textbf{100\%} & \textbf{100\%} \\
\textit{Handover Block} & 66\% & 57\% & 86\% & 73\% & \textbf{99\%} & 78\% & 95\% & 81\% & 97\% & \textbf{86\%} \\
\textit{Handover Mic} & 98\% & 97\% & 78\% & 63\% & 94\% & 96\% & \textbf{99\%} & \textbf{100\%} & 81\% & 92\% \\
\textit{Hanging Mug} & 18\% & 17\% & 38\% & 38\% & 40\% & 28\% & \textbf{58\%} & \textbf{62\%} & 47\% & 36\% \\
\textit{Lift Pot} & 96\% & 85\% & 96\% & 99\% & \textbf{100\%} & 99\% & \textbf{100\%} & \textbf{100\%} & 99\% & 97\% \\
\textit{Move Can Pot} & 51\% & 55\% & 34\% & 74\% & \textbf{94\%} & \textbf{97\%} & 90\% & 88\% & 88\% & 91\% \\
\textit{Move Pillbottle Pad} & 84\% & 61\% & 93\% & 96\% & 99\% & \textbf{99\%} & \textbf{100\%} & \textbf{99\%} & 99\% & 97\% \\
\textit{Move Playingcard Away} & 96\% & 84\% & \textbf{100\%} & 96\% & \textbf{100\%} & 99\% & \textbf{100\%} & \textbf{100\%} & 99\% & \textbf{100\%} \\
\textit{Move Stapler Pad} & 56\% & 42\% & 83\% & \textbf{85\%} & \textbf{91\%} & 79\% & 77\% & 64\% & 88\% & 76\% \\
\textit{Open Laptop} & 90\% & 96\% & 95\% & 91\% & 92\% & 94\% & \textbf{98\%} & \textbf{100\%} & 82\% & 78\% \\
\textit{Open Microwave} & 34\% & 77\% & \textbf{95\%} & \textbf{91\%} & 82\% & 86\% & 62\% & 45\% & 47\% & 48\% \\
\textit{Pick Diverse Bottles} & 81\% & 71\% & \textbf{90\%} & \textbf{91\%} & 89\% & 82\% & 80\% & 85\% & 89\% & 87\% \\
\textit{Pick Dual Bottles} & 93\% & 63\% & 96\% & 90\% & \textbf{100\%} & \textbf{99\%} & \textbf{100\%} & 96\% & 99\% & 98\% \\
\textit{Place A2B Left} & 87\% & 82\% & 88\% & 79\% & \textbf{97\%} & \textbf{93\%} & 95\% & \textbf{93\%} & 96\% & 88\% \\
\textit{Place A2B Right} & 87\% & 84\% & 91\% & 87\% & \textbf{97\%} & 95\% & 93\% & \textbf{99\%} & 95\% & 88\% \\
\textit{Place Bread Basket} & 77\% & 64\% & 91\% & 94\% & \textbf{97\%} & \textbf{95\%} & 91\% & 93\% & 89\% & 94\% \\
\textit{Place Bread Skillet} & 85\% & 66\% & 86\% & 83\% & \textbf{95\%} & 90\% & 90\% & \textbf{93\%} & 86\% & 86\% \\
\textit{Place Burger Fries} & 94\% & 87\% & 98\% & 98\% & 97\% & 95\% & 96\% & \textbf{99\%} & \textbf{99\%} & 97\% \\
\textit{Place Can Basket} & 62\% & 62\% & 81\% & 76\% & 81\% & \textbf{84\%} & 71\% & 69\% & \textbf{90\%} & 74\% \\
\textit{Place Cans Plasticbox} & 94\% & 84\% & 98\% & 94\% & \textbf{100\%} & \textbf{99\%} & 99\% & 96\% & 97\% & 98\% \\
\textit{Place Container Plate} & 99\% & 95\% & 98\% & 99\% & 99\% & 97\% & 96\% & \textbf{100\%} & \textbf{100\%} & 98\% \\
\textit{Place Dual Shoes} & 75\% & 75\% & 93\% & 87\% & \textbf{94\%} & \textbf{89\%} & \textbf{94\%} & 88\% & 43\% & 52\% \\
\textit{Place Empty Cup} & \textbf{100\%} & 99\% & 99\% & 98\% & \textbf{100\%} & \textbf{100\%} & \textbf{100\%} & \textbf{100\%} & \textbf{100\%} & 99\% \\
\textit{Place Fan} & 87\% & 85\% & 91\% & 87\% & \textbf{99\%} & 93\% & 96\% & \textbf{96\%} & 95\% & 92\% \\
\textit{Place Mouse Pad} & 60\% & 39\% & 66\% & 68\% & \textbf{93\%} & \textbf{96\%} & 83\% & 89\% & 89\% & 90\% \\
\textit{Place Object Basket} & 80\% & 76\% & 81\% & 87\% & \textbf{91\%} & \textbf{88\%} & 89\% & \textbf{88\%} & 87\% & 84\% \\
\textit{Place Object Scale} & 86\% & 80\% & 88\% & 85\% & \textbf{96\%} & 95\% & 90\% & \textbf{97\%} & 92\% & 81\% \\
\textit{Place Object Stand} & 91\% & 85\% & 98\% & \textbf{97\%} & \textbf{99\%} & 96\% & 90\% & 94\% & \textbf{99\%} & \textbf{97\%} \\
\textit{Place Phone Stand} & 81\% & 81\% & 87\% & 86\% & \textbf{97\%} & 97\% & \textbf{97\%} & \textbf{99\%} & 96\% & 94\% \\
\textit{Place Shoe} & 92\% & 93\% & 99\% & 97\% & 98\% & 98\% & 96\% & \textbf{99\%} & \textbf{100\%} & 94\% \\
\textit{Press Stapler} & 87\% & 83\% & 93\% & \textbf{98\%} & 85\% & 82\% & 90\% & 97\% & \textbf{96\%} & 97\% \\
\textit{Put Bottles Dustbin} & 84\% & 79\% & 81\% & 79\% & 87\% & \textbf{91\%} & \textbf{95\%} & 90\% & 93\% & 86\% \\
\textit{Put Object Cabinet} & 80\% & 79\% & 88\% & 71\% & 85\% & 87\% & \textbf{94\%} & \textbf{89\%} & 89\% & 78\% \\
\textit{Rotate QRcode} & 89\% & 87\% & 89\% & 73\% & \textbf{96\%} & \textbf{91\%} & 93\% & 89\% & 93\% & 87\% \\
\textit{Scan Object} & 72\% & 65\% & 67\% & 66\% & \textbf{96\%} & 91\% & 89\% & \textbf{92\%} & 84\% & 68\% \\
\textit{Shake Bottle Horizontally} & 99\% & 99\% & \textbf{100\%} & 98\% & \textbf{100\%} & 99\% & \textbf{100\%} & \textbf{100\%} & 98\% & 97\% \\
\textit{Shake Bottle} & 99\% & 97\% & \textbf{100\%} & 97\% & \textbf{100\%} & 97\% & \textbf{100\%} & \textbf{100\%} & \textbf{100\%} & 97\% \\
\textit{Stack Blocks Three} & 91\% & 76\% & 91\% & 95\% & \textbf{99\%} & \textbf{98\%} & 95\% & 97\% & 95\% & 83\% \\
\textit{Stack Blocks Two} & 97\% & \textbf{100\%} & \textbf{100\%} & 98\% & \textbf{100\%} & 98\% & \textbf{100\%} & \textbf{100\%} & \textbf{100\%} & 98\% \\
\textit{Stack Bowls Three} & 77\% & 71\% & 79\% & \textbf{87\%} & \textbf{86\%} & 83\% & 80\% & 81\% & 84\% & 81\% \\
\textit{Stack Bowls Two} & 95\% & 96\% & \textbf{98\%} & \textbf{98\%} & 94\% & \textbf{98\%} & 92\% & \textbf{98\%} & \textbf{98\%} & 96\% \\
\textit{Stamp Seal} & 79\% & 55\% & 93\% & 92\% & 96\% & 97\% & 90\% & 94\% & \textbf{97\%} & \textbf{99\%} \\
\textit{Turn Switch} & 62\% & 54\% & 84\% & 78\% & 44\% & 45\% & 61\% & 59\% & \textbf{91\%} & \textbf{80\%} \\
\midrule
\textbf{Average (\%)} & 82.74 & 76.76 & 88.66 & 87.02 & \textbf{92.90} & 91.50 & 91.88 & \textbf{91.78} & 90.50 & 87.46 \\
\bottomrule
\end{tabular}%
}
\end{table*}

\begin{table*}[t]
\centering
\caption{Per-task success rates (\%) across five robot embodiments on 31 shared RoboTwin 2.0 tasks under clean and randomized settings. Each task is evaluated over 25 episodes per embodiment and setting.}
\label{tab:robotwin_five_embodiments}
\small
\setlength{\tabcolsep}{3pt}
\resizebox{\textwidth}{!}{
\begin{tabular}{l cc cc cc cc cc}
\toprule
\multirow{2}{*}{\textbf{Simulation Task}} & \multicolumn{2}{c}{\textbf{Aloha}} & \multicolumn{2}{c}{\textbf{Franka}} & \multicolumn{2}{c}{\textbf{Piper}} & \multicolumn{2}{c}{\textbf{ARX-X5}} & \multicolumn{2}{c}{\textbf{UR5}} \\
\cmidrule(lr){2-3} \cmidrule(lr){4-5} \cmidrule(lr){6-7} \cmidrule(lr){8-9} \cmidrule(lr){10-11}
 & Clean & Random & Clean & Random & Clean & Random & Clean & Random & Clean & Random \\
\midrule
\textit{Beat Block Hammer} & 100\% & 100\% & 60\% & 64\% & 60\% & 76\% & 56\% & 68\% & 56\% & 48\% \\
\textit{Blocks Ranking RGB} & 0\% & 0\% & 0\% & 0\% & 4\% & 0\% & 0\% & 0\% & 0\% & 0\% \\
\textit{Blocks Ranking Size} & 64\% & 64\% & 68\% & 56\% & 80\% & 84\% & 52\% & 60\% & 12\% & 8\% \\
\textit{Grab Roller} & 100\% & 96\% & 96\% & 96\% & 100\% & 100\% & 100\% & 100\% & 32\% & 48\% \\
\textit{Handover Mic} & 80\% & 80\% & 40\% & 28\% & 96\% & 100\% & 100\% & 96\% & 48\% & 40\% \\
\textit{Lift Pot} & 100\% & 96\% & 72\% & 68\% & 96\% & 100\% & 96\% & 96\% & 0\% & 0\% \\
\textit{Move Can Pot} & 72\% & 72\% & 48\% & 68\% & 68\% & 60\% & 84\% & 84\% & 24\% & 8\% \\
\textit{Move Pillbottle Pad} & 84\% & 88\% & 68\% & 64\% & 100\% & 92\% & 96\% & 100\% & 4\% & 0\% \\
\textit{Move Playingcard Away} & 96\% & 100\% & 96\% & 100\% & 100\% & 100\% & 100\% & 100\% & 88\% & 88\% \\
\textit{Move Stapler Pad} & 92\% & 96\% & 52\% & 32\% & 84\% & 80\% & 88\% & 92\% & 16\% & 24\% \\
\textit{Open Laptop} & 96\% & 96\% & 56\% & 64\% & 60\% & 80\% & 96\% & 96\% & 48\% & 64\% \\
\textit{Open Microwave} & 28\% & 36\% & 16\% & 36\% & 16\% & 24\% & 12\% & 20\% & 20\% & 28\% \\
\textit{Place A2B Left} & 92\% & 84\% & 92\% & 72\% & 96\% & 84\% & 92\% & 76\% & 20\% & 12\% \\
\textit{Place A2B Right} & 76\% & 84\% & 80\% & 92\% & 84\% & 76\% & 96\% & 92\% & 20\% & 20\% \\
\textit{Place Bread Basket} & 84\% & 68\% & 76\% & 44\% & 64\% & 64\% & 80\% & 76\% & 56\% & 40\% \\
\textit{Place Burger Fries} & 96\% & 100\% & 84\% & 80\% & 92\% & 100\% & 96\% & 96\% & 20\% & 24\% \\
\textit{Place Container Plate} & 100\% & 100\% & 100\% & 100\% & 100\% & 100\% & 100\% & 100\% & 68\% & 48\% \\
\textit{Place Dual Shoes} & 64\% & 64\% & 16\% & 40\% & 76\% & 96\% & 48\% & 36\% & 4\% & 0\% \\
\textit{Place Empty Cup} & 100\% & 100\% & 100\% & 96\% & 96\% & 68\% & 92\% & 52\% & 68\% & 36\% \\
\textit{Place Mouse Pad} & 80\% & 88\% & 84\% & 68\% & 92\% & 100\% & 100\% & 96\% & 4\% & 8\% \\
\textit{Place Object Scale} & 84\% & 88\% & 84\% & 72\% & 80\% & 100\% & 80\% & 96\% & 44\% & 32\% \\
\textit{Place Object Stand} & 100\% & 100\% & 92\% & 84\% & 96\% & 96\% & 96\% & 88\% & 16\% & 20\% \\
\textit{Place Phone Stand} & 96\% & 96\% & 8\% & 16\% & 92\% & 88\% & 100\% & 92\% & 12\% & 0\% \\
\textit{Place Shoe} & 96\% & 100\% & 4\% & 4\% & 88\% & 92\% & 92\% & 96\% & 4\% & 8\% \\
\textit{Press Stapler} & 92\% & 100\% & 100\% & 96\% & 100\% & 92\% & 100\% & 92\% & 96\% & 84\% \\
\textit{Shake Bottle Horizontally} & 100\% & 100\% & 68\% & 52\% & 100\% & 100\% & 100\% & 100\% & 36\% & 44\% \\
\textit{Shake Bottle} & 100\% & 96\% & 80\% & 68\% & 100\% & 92\% & 100\% & 92\% & 48\% & 68\% \\
\textit{Stack Blocks Two} & 16\% & 4\% & 16\% & 20\% & 12\% & 16\% & 4\% & 8\% & 8\% & 16\% \\
\textit{Stack Bowls Two} & 92\% & 92\% & 88\% & 92\% & 100\% & 100\% & 96\% & 96\% & 60\% & 52\% \\
\textit{Stamp Seal} & 76\% & 88\% & 56\% & 64\% & 100\% & 92\% & 100\% & 100\% & 8\% & 28\% \\
\textit{Turn Switch} & 84\% & 80\% & 100\% & 100\% & 96\% & 100\% & 96\% & 92\% & 48\% & 36\% \\
\midrule
\textbf{Average (\%)} & 81.94 & 82.45 & 64.52 & 62.45 & 81.55 & 82.32 & 82.19 & 80.26 & 31.87 & 30.06 \\
\bottomrule
\end{tabular}
}
\end{table*}

\end{document}